\documentclass[10pt]{article} 

\usepackage[utf8]{inputenc} 
\usepackage[T1]{fontenc}    
\usepackage{url}            
\usepackage{booktabs}       
\usepackage{amsfonts}       
\usepackage{nicefrac}       
\usepackage{microtype}      
\usepackage{xcolor}         

\usepackage[utf8]{inputenc}
\usepackage{amsmath,amssymb,amsthm,mathtools}
\usepackage[a4paper,margin=1in]{geometry}
\usepackage[numbers,sort&compress]{natbib}
\usepackage{slashed}
\usepackage{braket}
\usepackage{siunitx}       
\usepackage{booktabs}     
\usepackage{tikz}     
\usepackage{enumitem}
\usepackage{tabularx}   
\usepackage{array}  

\usepackage{longtable}
\usepackage{makecell}

\usepackage[hidelinks]{hyperref} 
\usepackage{algorithm}
\usepackage{algorithmic} 

\usetikzlibrary{arrows.meta,positioning,fit,shapes,calc}

\usetikzlibrary{arrows.meta,positioning,fit,shapes,shapes.geometric,calc}

\theoremstyle{definition}

\theoremstyle{plain}

\theoremstyle{remark}

\title{When Rule Violations Are Rare: Chimera Training for Logical Anomaly Detection}

\author{1\textsuperscript{st} Alejandro Asc\'arate\thanks{
\small School of Electrical Engineering and Robotics, Faculty of Engineering,\\
\small $-\;\,\,\,\text{Queensland}$ University of Technology, Brisbane, Queensland, Australia} \\
a.ascaratecastro@hdr.qut.edu.au
\and
2\textsuperscript{nd}  L\'eo Lebrat$^{*}$ \\
leo.lebrat@qut.edu.au
\and
3\textsuperscript{rd} Rodrigo Santa Cruz$^{*}$ \\
rodrigo.santacruz@qut.edu.au
\and
4\textsuperscript{th} Clinton Fookes$^{*}$ \\
c.fookes@qut.edu.au
\and
5\textsuperscript{th} Olivier Salvado$^{*}$ \\
olivier.salvado@qut.edu.au
}

\begin{document}

\maketitle

\begin{abstract}
Many practical anomalies are not merely rare inputs, but violations of semantic constraints: objects co-occur in structured ways, actions imply preconditions, and events satisfy temporal or relational regularities. We study anomaly detection in this setting, where constraints are given as logical rules over learned visual concepts, but real rule violations are rare or absent during training. We propose a neural rule evaluator that compiles each constraint into a directed acyclic graph and learns feature-aware subtree MLP gates for its internal logical operators. Each gate maps child features and edge-level negations to a parent representation and a rule-satisfaction probability, with intermediate supervision obtained from exact Boolean propagation over ground-truth concept labels. The key difficulty is that same-image training data often provide insufficient coverage of informative truth configurations and also allow shortcut solutions. To address this, we introduce chimera training: an operand-level counterfactual construction at the feature level. Instead of mixing input images, we concatenate subtree features from different samples; each operand keeps the hard truth label of the sample it came from, and the chimera target is obtained by applying the node’s logical operator to those inherited labels. This supplies supervised logical counterexamples without requiring real anomalous images. Across CLEVRER, OpenImages, and VidOR, the resulting evaluator improves rule-level anomaly AUROC over independent-events and same-image semantic-training baselines, especially for compositional and relational rules. The method yields both scalar anomaly scores and rule-level attributions.
\end{abstract}

\section{Introduction}

\begin{figure}[h!]
    \centering
    \includegraphics[width=1\linewidth]{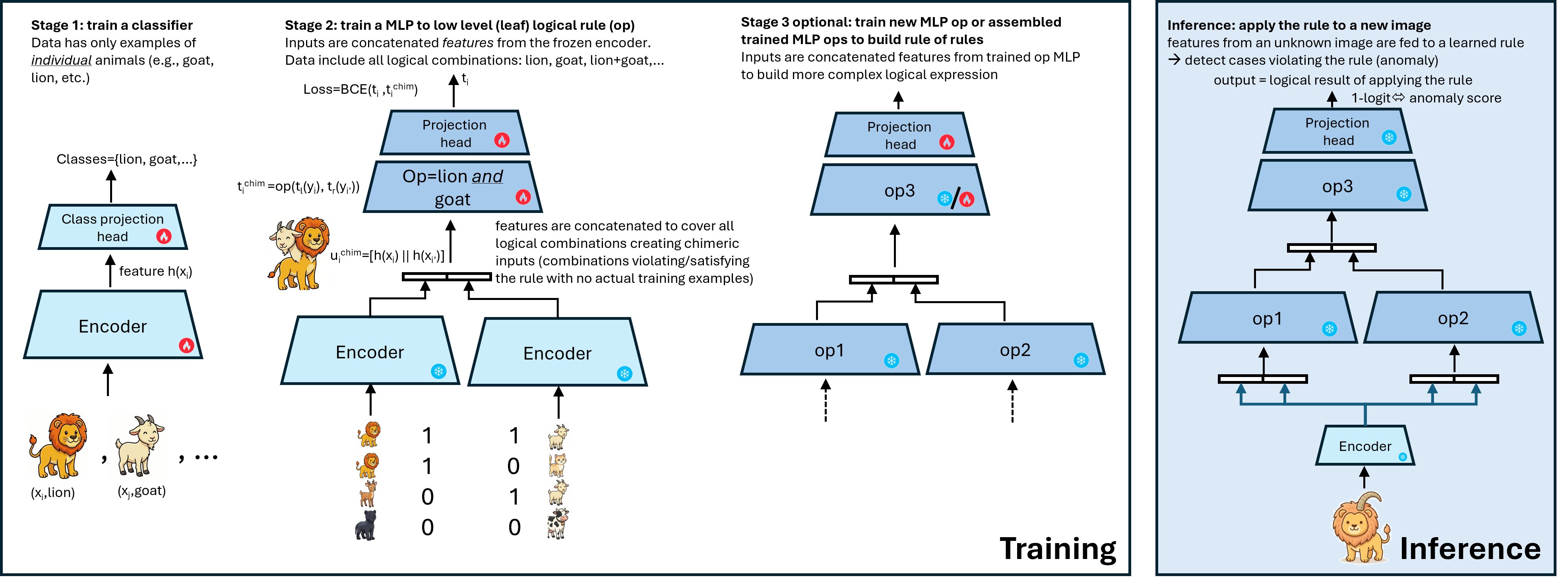}
    \caption{Training and inference of our proposed method. In the first stage a classifier is trained. The head is discarded while the features of the backbone are used next. Each node logical rule needs to be learned using a specific model (stage 2). The backbones of the node rules can be used to learn more complex expressions (stage 3). Once a logical expression has been learned (e.g. a logical `$\mathrm{and}$'), it can be frozen and assembled to form more complex rules (alternate stage 3). During inference (stage 4) the frozen ops and classifier backbone are applied to a new image to apply a rule. The output can then be used for anomaly detection or logical inference from the image attributes.}
    \label{chims}
\end{figure}

Detecting anomalies from a dataset is often framed as statistical detection: detecting outliers, samples outside the data distribution (out of distribution, OOD) \citep{Chandola2009Survey,Hendrycks2017Baseline,Ruff2021UnifyingReview}. Instead, we are concerned with detecting anomalous samples on the basis of those samples breaking known rules usually satisfied by the initial distribution. This Logical Anomaly Detection approach usually requires to detect and/or learn all the possible logical cases. Our new approach in this paper, describes a method that does not require identifying all the logical cases and can thus detect if a rule is broken when the training dataset contains only samples that are consistent with the rule (missing the anomalous cases). This is a key distinguishing aspect of the anomaly detection problem. In comparison, a fully supervised binary classification would require access to training samples of anomalies, which are by definition rare and usually not accessible in large quantities for training. 

\begin{figure}[!h]
    \centering
    \includegraphics[width=1\linewidth]{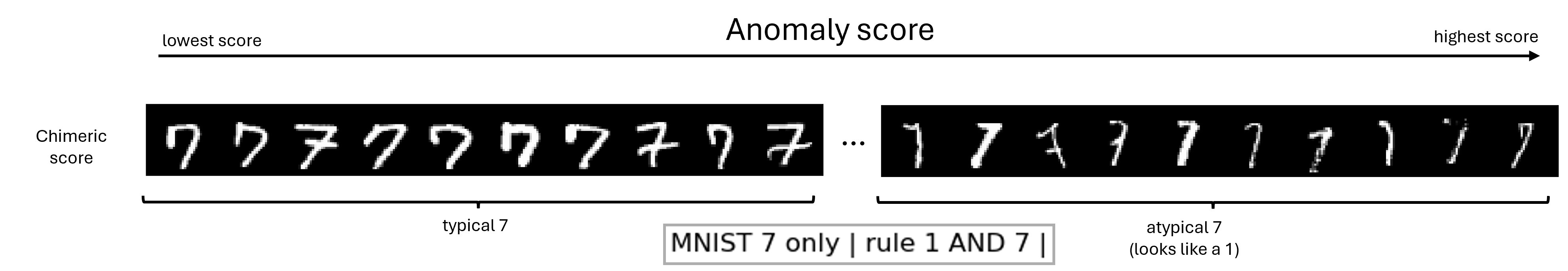}
    \caption{Score-sorted MNIST test images for the rule \(1 \wedge 7\), restricted to true digit-\(7\) samples. Images are ordered by increasing learned conjunction score \(\widehat{P}(1 \wedge 7)\); high-scoring examples correspond to atypical \(7\)'s whose stroke geometry also activates evidence for digit \(1\).}
    \label{fig:mnistchims2}
\end{figure}

\begin{figure}[!h]
    \centering
    \includegraphics[width=1\linewidth]{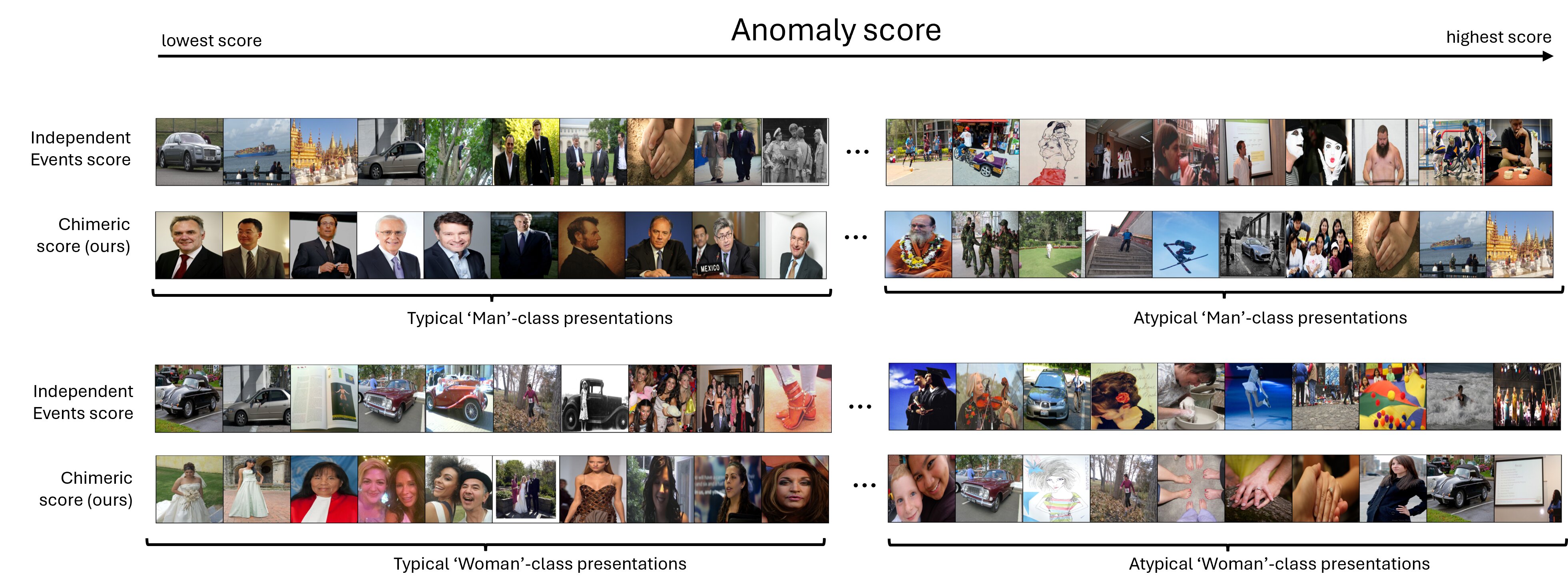}
    \caption{\textbf{Qualitative visualization of results for the OpenImages contradiction rule} ${A \Leftrightarrow \neg A}$. In this experiment, the anomaly score is the model output for the contradiction rule itself (not $1-p$). Independent Events calculates it only on the basis of the initial classifier's logits, whereas the chimera training version trains an MLP gate for the rule while additionally introducing synthetic contradictory examples at the feature level that cannot occur in real data. For a fixed test class, we sort samples by anomaly score and show the 10 smallest on the left and the 10 largest on the right. Although all shown images share the same dataset label, the high-score samples are visually more distorted and less prototypical. Chimera produces a visibly cleaner separation, tending to place more normal-looking instances on the left and more abnormal-looking ones on the right, which suggests that the synthetic contradictory supervision helps the evaluator detect within-class visual abnormality more effectively. Upper panel, rule is `${\text{man} \Leftrightarrow \neg \,\text{man}}$'; lower panel, rule is `${\text{woman} \Leftrightarrow \neg \,\text{woman}}$'.}
    \label{fig:OIviz}
\end{figure}

Using the classic simple example of digits classification in MNIST, our method allows detecting all the labeled ``7'' that look like ``1'' (see Fig.\ref{fig:mnistchims2}). When applied to natural images, one could identify all the atypical presentations in images labeled ``(wo)man'' as shown in Fig.\ref{fig:OIviz} (from the OpenImages dataset). More complex `rules' can be learned as long as they can be expressed as functions of attributes that can be estimated using an appropriate model. We show examples of \textit{causal} rules from videos of basic moving shapes such as ``$\text{collide}(\text{shape=sphere},\text{color=red}) \implies \text{collide\_before\_half\_of\_video}$'' (using the CLEVRER dataset), and realistic rules on complex and changing scenes in videos such as ``$\text{obj:baby} \implies (\text{rel:baby-in\_front\_of-adult} \wedge \text{rel:adult-watch-baby})$'' (using the VidOR dataset).

Thus, many anomaly scenarios are better characterized not merely by ``rarity'' in pixel (or even latent) space, but by violations of \emph{domain constraints}: objects co-occur in structured ways; actions imply preconditions; and relationships satisfy logical regularities. If such constraints are available (hand-written, mined, or curated), they can serve as a semantically meaningful interface for detection: an input is anomalous if it contradicts one or more constraints. 


However, integrating constraint evaluation with high-dimensional perception is nontrivial. Fully symbolic pipelines require brittle perception outputs; fully neural pipelines often re-learn constraints implicitly and entangle them with spurious cues.

This paper proposes a neuro-symbolic anomaly detection framework that treats constraints as \emph{explicit computation graphs} (e.g., binary \textit{trees}) and learns reusable \emph{neural operators} (which we call `gates') that implement logical composition. The method will be referred to as `Neural Evaluator' in the rest of the paper (see Fig.\ref{chims}).

A central challenge is preventing gates from collapsing into shortcut classifiers for the entire rule (e.g., recognizing an anomaly template directly from image features). We address this via \emph{`chimera'\footnote{\href{https://en.wikipedia.org/wiki/Chimera_(mythology)}{Wikipedia-Chimera\_(mythology)}.} negative training} (see Fig.\ref{chims}, also Sec.\ref{sec:method}). This construction is conceptually related to mixup-style interventions \citep{Zhang2018Mixup,Yun2019CutMix}, but operates at the level of \emph{subtree operands} rather than raw pixels or labels. Empirically, this encourages gates to behave as compositional operators and improves transfer of learned subtrees across constraints. Furthermore, for many rules, informative counterexamples are exceedingly rare under the natural data distribution (e.g., implication violations require antecedent true and consequent false). Consequently, training a global predictor on real observed samples only yields degenerate solutions (see table \ref{tab:rule_aggregate}, `SEM').

\subsection{Scope, task definition, and non-goals}
\label{sec:scope}

Because the setting considered here lies at the intersection of anomaly
detection, neuro-symbolic reasoning, and compositional neural models, it is
important to distinguish the learning problem studied in this work from several
related but different problems.

\paragraph{The rule is supplied, not learned.}
The logical rule $R$ is an exogenous input to the system: it is specified in
advance, compiled into a fixed computation graph, and its Boolean semantics are
known. The learning problem is to estimate the truth of this specified rule,
and of its subexpressions, from perceptual evidence. We therefore do not study
rule induction or zero-shot generalization to previously unseen logical
expressions. In particular, ``compositionality'' in the present work refers to
the bottom-up evaluation of the subexpressions of a supplied rule and to the
possible reuse of already trained subtrees when the same subexpression occurs
elsewhere. Learning a general algebra that can be applied to arbitrary new
rules is a separate problem.

\paragraph{The setting is zero-violation anomaly detection, not few-shot
learning.}
The target regime is one in which no observed training sample is required to
violate the rule. Real rule violations may be completely absent from the
nominal training set. The problem is therefore not to adapt from one or a few
labelled anomalous examples, as in few-shot classification or few-shot anomaly
detection. Chimera training instead constructs the missing logical supervision
using only nominal samples: operands extracted from different samples are
combined, and the exact target of the resulting configuration follows from the
known Boolean connective. Methods that assume access to one or more labelled
violations consequently address a different supervision regime.

\paragraph{The neural evaluator does not assume independent concepts.}
Statistical independence between leaf events is assumed only by the
\textsc{IndepProb} baseline, which computes rule probabilities by factorizing
the predicted marginal concept probabilities. The proposed neural evaluator
does not make this assumption: its gates operate on learned encoder features
and are trained against exact Boolean node targets. The comparison between
approximately separable and strongly relational or temporal settings is included
precisely to test when the independence approximation becomes inadequate, not
because independence is an assumption of the proposed method.

\paragraph{Ground-truth concepts define supervision, not test-time inputs.}
Ground-truth concept annotations are used to construct exact Boolean targets
during training and to define the reference truth value of a rule for
performance testing purposes once trained. They are not supplied to the neural evaluator at inference time while doing the inference itself.
Test-time rule evaluation starts from the learned visual leaf bank and its
encoder features, and therefore includes errors introduced by imperfect
perception. VidOR provides a particularly stringent natural example: its sparse
relation vocabulary yields very low leaf-level macro average precision, while
the downstream neural evaluator nevertheless retains substantial rule-level
discrimination. The method therefore does not assume near-perfect concept
prediction at inference at all, in fact quite the contrary.

\paragraph{Chimeras are semantic training configurations, not generated
anomalous images.}
A chimera is not intended to approximate a realistic anomalous image, and no
image is synthesized. The intervention takes place at the operand-feature level.
Its purpose is to expose a neural gate to a Boolean configuration that is absent
from the nominal training distribution. The validity of its training target
comes from the exact semantics of the corresponding connective, rather than
from the visual realism of the mixed representation. At test time, the
evaluator is applied only to ordinary same-sample operands. Chimera training is
therefore fundamentally different from image-space augmentation schemes whose
synthetic inputs themselves must be assigned an approximate semantic label.

\paragraph{Cross-sample training and same-sample inference are deliberate.}
During chimera training, operands are intentionally taken from different
nominal samples in order to break the empirical coupling that prevents some
truth configurations from appearing naturally. During inference, all operands
again arise from the same image or video. This asymmetry is the intervention
being studied rather than an accidental train--test mismatch: all reported
anomaly-detection results directly test transfer from cross-sample training to
natural same-sample inference. We additionally find that augmenting chimera
training with ordinary same-sample pairs changes performance only marginally,
indicating that the critical source of supervision is the missing
cross-operand truth configurations rather than exposure to the test-time
pairing pattern itself.

\paragraph{Same-image training and monolithic models are controlled
ablations.}
The same-image semantic-training variant and the monolithic models are included
to isolate the source of the observed improvement; they are not intended as
stand-alone state-of-the-art anomaly-detection methods. Same-image training
retains the rule representation, perceptual features, Boolean targets, and
evaluation protocol while removing only the chimera intervention. Similarly,
the monolithic comparison retains chimera supervision while removing the
node-local rule decomposition. Together these ablations distinguish the effect
of counterfactual truth-configuration coverage from that of the modular
architecture.

\paragraph{Predictive gain and modular structure play different roles.}
The chimera-trained monolithic ablation performs close to the full evaluator,
showing that the principal improvement in anomaly-detection accuracy comes
from chimera supervision rather than from the DAG architecture itself. We do
not therefore claim a large additional AUROC gain from modularity. The DAG has
a different role: it makes the supplied logical decomposition explicit, exposes
intermediate subformula scores, and permits a previously trained subtree to be
reused when the same symbolic structure and feature lineage recur. The
experimental conclusions should consequently be separated: chimera training
addresses the statistical problem of missing logical support, whereas the DAG
provides the structured interface through which rule evaluation is decomposed.

\paragraph{Anomalies are rule-relative, and attributions are not causal proofs.}
An ``anomaly'' in the present setting is a sample that falsifies a supplied
domain rule; it is not required to belong to an independently defined generic
anomaly class. A sample may therefore be anomalous with respect to one
constraint and nominal with respect to another. Likewise, the rule and
subformula scores exposed by the evaluator are semantic attributions: they
identify which supplied constraints are estimated to be violated. They are not
claimed to constitute causal explanations of the observation or formal proofs
that the learned neural gates satisfy Boolean identities for every possible
feature vector. Formal verification, causal explanation, and automatic rule
discovery are separate problems outside the scope of this work.

\paragraph{Contributions.}
\begin{itemize}
    \item We propose \emph{subtree gates}, a bottom-up node-local-learned evaluator that composes concept-conditioned features into truth probabilities under hard Boolean supervision.
    \item We introduce \emph{chimera negative training} to enforce operator-level compositionality, reduce shortcut learning in rule evaluators, and to operate in cases where real counterfactuals to the rules are completely absent in the training data (most real applications in anomaly detection).
    \item We demonstrate effectiveness on structured image/video benchmarks and real-world data (CLEVRER, OpenImages, VidOR).
\end{itemize}

\section{Related Work}
\label{sec:related}

\paragraph{Deep anomaly and OOD detection}
A large body of work scores anomalies using density or reconstruction surrogates (e.g., autoencoders/VAEs), feature-distance criteria, or uncertainty estimates \citep{Chandola2009Survey,Ruff2021UnifyingReview}. In modern deep OOD detection, common baselines include softmax confidence \citep{Hendrycks2017Baseline}, input perturbation and temperature scaling \citep{Liang2018ODIN,Guo2017Calibration}, and feature-space detectors such as Mahalanobis distances \citep{Lee2018Mahalanobis}. Energy-based scoring has also emerged as a unifying view for some classification models \citep{Liu2020Energy}. These approaches typically provide a scalar score with limited semantic attribution: they rarely explain \emph{which} structured expectation is violated.

\paragraph{Logical anomaly detection.}
A more directly related line of work studies anomalies whose abnormality is
defined by semantic composition or global consistency rather than by a local
visual defect. MVTec LOCO and subsequent logical-anomaly methods establish
this distinction explicitly in industrial inspection, while methods such as
PSAD reason about abnormal part configurations and LogicAL constructs
synthetic logical anomalies to improve anomaly localization
\cite{Bergmann2022LogicalAnomaly,Kim2024PSAD,Zhao2024LogicAL}. These works are important precedents for the
\emph{logical anomaly} problem itself, and our contribution should not be read
as introducing that problem formulation or the general idea of synthesizing
anomalous configurations.

The distinction lies in the object being synthesized and in the supervision
that results from it. Chimera training does not construct a visually realistic
anomalous image. Given an externally supplied Boolean rule, it combines
sample-level subtree operands in feature space so as to instantiate truth
configurations that are absent from the nominal data, and assigns them exact
targets from the known connective. Its purpose is therefore to make a neural
evaluator of the supplied rule learnable in the strict zero-observed-violation
regime. We view image-space logical-anomaly synthesis and operand-level
logical supervision as complementary solutions to different parts of the
logical-anomaly-detection problem.

\paragraph{Semantic and constraint-aware anomaly detection}
A complementary line leverages structure, constraints, or knowledge to detect implausible samples. Constraint-based and logic-guided learning often imposes penalties for rule violations or encourages outputs to satisfy known relations \citep{Hu2016Harnessing,Xu2018SemanticLoss}. Related ideas appear in weakly-supervised and knowledge-driven settings, where symbolic constraints regularize predictors without requiring full labels \citep{Ganchev2010PR}. While effective, many methods treat constraints as a global regularizer and do not explicitly construct reusable \emph{modules} implementing logical composition that can be transferred across many rules.

\paragraph{Neuro-symbolic reasoning and differentiable logic}
Neuro-symbolic methods aim to combine sub-symbolic perception with symbolic reasoning, including differentiable logical frameworks and probabilistic logic programming \citep{Garcez2019NeuroSymbolic,Manhaeve2018DeepProbLog,Donadello2017LTN}. In vision-and-language and synthetic reasoning benchmarks (e.g., CLEVR), neural module networks and related compositional models assemble learned operators according to an explicit program or graph structure \citep{Andreas2016NMN,Johnson2017Inferring}. These works motivate our design choice of compiling constraints into explicit computation graphs. However, much of this literature targets question answering or program execution rather than anomaly detection; and many methods learn operators end-to-end without a mechanism to (i) supervise intermediate truth semantics from ground-truth concepts, (ii) prevent shortcut learning at internal nodes, (iii) deal with highly unbalanced training data for a standard supervised fit, and, (iv) reuse learned subtrees safely across rule sets and runs.

\paragraph{Concept bottlenecks and interpretable interfaces}
Concept bottleneck models (CBMs) and related ``predict-then-reason'' pipelines provide an interpretable intermediate representation through human-aligned concepts \citep{Koh2020CBM}. Our leaf concept bank is similar in spirit: it exposes a semantically meaningful interface for downstream reasoning. Unlike standard CBM pipelines that apply a fixed symbolic reasoner (or a shallow classifier) atop concept predictions, we learn a \emph{structured evaluator} that maps concept-conditioned features through a constraint DAG, producing per-rule satisfaction probabilities and anomaly attributions.

\paragraph{Compositional regularization and counterfactual mixing}
Data mixing strategies such as mixup and CutMix improve robustness by constructing interpolated or patched examples \citep{Zhang2018Mixup,Yun2019CutMix}. Our \emph{chimera negative training} is related in the sense of creating counterfactual combinations, but differs in locus and supervision: we mix \emph{subtree operands} (child features) rather than raw inputs, and supervise targets using exact Boolean semantics computed from the corresponding hard child truths. This pushes internal gates to implement the intended connective rather than overfitting to global visual templates of a particular rule.

\paragraph{Neural Algebra of Classifiers and compositional rule learning.}
Neural Algebra of Classifiers (NAC; \citep{Cruz2018NeuralClassifiers}) and related methods
\citep{Misra_2017_CVPR,Nagarajan_2018_ECCV,Yang_2020_CVPR,Li_2021_WACV} are close architectural
precedents in that they learn neural mechanisms associated with logical
composition. Their principal learning problem, however, is different from the
one considered here. NAC learns operations over classifier representations so
that classifiers for previously unseen composed expressions can be synthesized
from classifiers learned during training. The principal generalization axis is
therefore from a set of \emph{fully supervised} expressions to new expressions.

Here, by contrast, the target rule is supplied in advance and its semantics are
fixed. We do not ask the system to infer how to represent or evaluate an
arbitrary unseen rule. The difficulty is instead that the natural training
distribution may contain no examples of one or more truth assignments required
to learn the evaluator of the supplied rule. Chimera training addresses this
\emph{support problem}: it creates the missing operand configurations from
nominal samples while retaining exact Boolean supervision. Thus, NAC-style
parameter composition and Chimera-style operand composition should not be
interpreted as competing solutions to the same learning objective. The former
primarily addresses generalization across expressions; the latter addresses
evaluation of a specified expression under missing empirical truth-table
support.

\begin{table}[t]
\centering
\caption{Conceptual distinction between NAC-style compositional learning and
the problem studied here.}
\label{tab:nac_scope}
\small
\begin{tabular}{p{0.25\linewidth}p{0.33\linewidth}p{0.33\linewidth}}
\toprule
 & \textbf{NAC-style learning} & \textbf{Chimera evaluator} \\
\midrule
Rule at evaluation
& May be previously unseen
& Supplied in advance \\

Primary problem
& Generalization across composed expressions
& Evaluation under missing truth-assignment support \\

Object being composed
& Classifier/parameter representations
& Sample-level subtree operands \\

Source of supervision
& Training expressions and their labels
& Concept labels plus exact semantics of the supplied rule \\

Real violating examples
& Not the defining issue
& May be completely absent \\

Main claim
& Learn an algebra reusable on new expressions
& Learn a specified rule evaluator without observed violations \\
\bottomrule
\end{tabular}
\end{table}



\section{Method}
\label{sec:method}
\subsection{Problem set up, training, and inference}
\paragraph{Problem set up}We consider a multi-label dataset
\begin{equation}
\mathcal{D}=\{(x_i,y_i)\}_{i=1}^M,\qquad y_i\in\{0,1\}^N,
\end{equation}
where $y_{i,c}=1$ indicates that concept $c$ is present in $x_i$. We are given rules
$\{\mathcal{R}_r\}_{r=1}^R$, each compiled into a directed acyclic graph
$G_r=(V_r,E_r)$ whose leaves are concept IDs and whose internal nodes are logical
operators in $\{\mathrm{IFF},\mathrm{IMPLIES},\mathrm{AND},\mathrm{OR}\}$. Edges may carry
negation flags. The task is to output an anomaly score $s(x)\in[0,1]$ together with
per-rule violation scores $\{s_r(x)\}$.

\paragraph{Training}A leaf concept bank first produces
\begin{equation}
z=E_\phi(x)\in\mathbb{R}^F,\qquad
\ell(x)\in\mathbb{R}^N,\qquad
p(x)=\sigma(\ell(x))\in(0,1)^N .
\end{equation}
The encoder feature $z$ is used as the evidence carrier for neural rule evaluation, while
$p(x)$ is used for scoring.

Each rule graph stores node attributes identifying leaves, concept IDs, and operator codes,
and edge attributes encoding negation and, for implications, operand order. Given hard
concept labels $y$, exact Boolean semantics are propagated bottom-up through the graph to
obtain node-level targets $t_v(y)\in\{0,1\}$ for every internal node, not only the root.

For each internal node $v$ with children $c_1,\dots,c_{a_v}$, we assign a learned subtree gate
$g_{\theta_v}$. With child features $h_{c_j}\in\mathbb{R}^F$ and negation indicators
$b_j\in\{0,1\}$, define
\begin{align}
u_v &= [h_{c_1}\|b_1\|\cdots\|h_{c_{a_v}}\|b_{a_v}],\\
h_v &= f_{\theta_v}(u_v),\\
\hat t_v &= \sigma(w_v^\top h_v+\beta_v).
\end{align}
Leaves are initialized with $h_v\leftarrow z=E_\phi(x)$. Thus, rule structure determines
which concept role each copy of $z$ plays, while gates learn concept- and operator-specific
composition in feature space.

Training is performed bottom-up by depth. For depth level
$\mathcal{V}_d=\{v:\mathrm{depth}(v)=d\}$, lower-depth gates are frozen and used to produce
child features. Each gate at level $d$ is then trained with node-wise binary cross-entropy,
\begin{equation}
\mathcal{L}_v
=
\frac{1}{B}\sum_{i=1}^B
\mathrm{BCE}\big(\hat t_v(x_i),t_v(y_i)\big).
\end{equation}
This internal supervision forces local logical composition rather than a monolithic root-level
rule classifier.

To prevent shortcut learning, we use chimera training. For a binary node $v$ with children
$(\ell,r)$, choose a permutation $\pi$ with $\pi(i)\neq i$ and form mixed operands
\begin{equation}
(u_v^{\mathrm{chim}})_i
=
[h_\ell(x_i)\|b_\ell\|h_r(x_{\pi(i)})\|b_r].
\end{equation}
The target is computed by the intended Boolean operator:
\begin{equation}
(t_v^{\mathrm{chim}})_i
=
\mathrm{op}(v)\big(t_\ell(y_i),t_r(y_{\pi(i)})\big).
\end{equation}
Thus, counterfactual mixing occurs at the operand level, giving informative truth assignments
even when some rule outcomes are rare or absent in the observed data.

Finally, to scale across many rules, trained gates are reused through lineage-aware caching.
A subtree key records the symbolic subtree, edge negations, operator order, gate architecture,
feature dimension, and a fingerprint of the upstream encoder. Hence a cached gate is reused
only when both the logical structure and the feature representation match.


\paragraph{Inference and anomaly scoring}
\label{sec:anomaly_scoring}
For each rule $r$, we evaluate the root satisfaction probability $\hat{t}^{(r)}_{\mathrm{root}}(x)\in(0,1)$ by bottom-up propagation through the trained/cached gates on $G_r$ (see Algorithm \ref{alg:predict-root-fn}), where again we use $\texttt{dgl.topological\_nodes\_generator}$ to great advantage.

\paragraph{Per-rule violation.}
We define a basic violation score (see Algorithm \ref{alg:inference}) 
\begin{equation}
v_r(x) = 1 - \hat{t}^{(r)}_{\mathrm{root}}(x).
\end{equation}

Because the scoring is rule-decomposable, the detector naturally provides semantic attributions: the top-$k$ rules with the largest $s_r(x)$ explain the anomaly.

\subsection{Why implications are particularly suited for anomaly detection.}
Among Boolean connectives, the implication $\;A \Rightarrow B\;$ is especially well-matched to anomaly detection because it's \textit{non-symmetric} and its truth table has a \emph{single} and \textit{non-trivial} falsifying configuration: it is false only when the antecedent holds but the consequent does not, i.e.\ $(A=1,\,B=0)$. Consequently, each implication directly defines a sharp notion of ``violation'' (our anomaly signal) that is both sparse and interpretable: an anomaly corresponds precisely to the presence of the contextual precondition $A$ together with the absence of the expected outcome $B$. Equivalently, the implication acts as a \emph{context-gated} detector: when the antecedent is not present ($A=0$), the rule is vacuously satisfied and produces no alarm (it ignores the sample outside its intended regime of applicability), whereas when the antecedent is present ($A=1$), the rule activates and flags an anomaly exactly when the consequent is absent ($B=0$) in that specific context. This ``activate-on-context, trigger-on-violation'' behavior is precisely what we want in realistic perceptual settings, where we aim to avoid spurious alarms on irrelevant inputs while reliably detecting context-specific inconsistencies.

\section{Experiments}
\label{sec:experiments}

\subsection{Datasets and concept vocabularies}
We evaluate on three vision benchmarks where (i) a multi-label concept inventory is available (or can be induced), and (ii) logical constraints over these concepts are meaningful.

\textbf{CLEVR (images).} We use CLEVR images and their ground-truth scene annotations \cite{Johnson2017CLEVR}. \textbf{CLEVRER (videos).} We use CLEVRER and its structured annotations \cite{Yi2020CLEVRER}. \textbf{Open Images.}
We use Open Images V4 annotations \cite{Kuznetsova2018OpenImagesV4}. \textbf{VidOR.}
We evaluate on the VidOR (Video Object Relation) dataset~\cite{shang2019annotating}.


\subsection{Evaluation tasks and metrics}
\paragraph{Consistency anomaly detection.}
We define a binary anomaly label from ground-truth concepts and rules:
\[
y_{\text{anom}}(x)=\mathbb{I}\Big[\min_{r\le R}\; \text{Truth}_r(x)=0\Big],
\]
i.e., an example is anomalous iff it violates \emph{at least one} constraint under hard boolean evaluation.
We report AUROC and (when useful) FPR@95TPR for anomaly detection.

\paragraph{Concept prediction quality.}
We report macro AUROC and macro AUPRC over the $K$ concept heads on the held-out split, since rule performance depends on leaf quality.

\paragraph{Rule complexity is constrained by the semantic support of the benchmark.}
Evaluation on larger and richer rule systems is an important
direction, but simply increasing formula depth or the number of rules does not
necessarily constitute a stronger test.  A rule can be evaluated meaningfully
only when the dataset annotations support sufficiently many non-degenerate
satisfying and falsifying instances on the held-out split.  The semantic
complexity of the available benchmarks therefore places an upper bound on the
rule complexity that can be validated empirically.  One can trivially generate
deeper Boolean expressions over the same concepts, but if their truth values
are determined by redundant subclauses, or if they are almost always true or
false, such expressions test graph depth rather than a new reasoning ability.

This distinction is especially relevant here: Chimera training removes the
need for naturally occurring rule violations in the \emph{training} set, but
held-out violations are still required to test whether the learned
evaluator transfers to genuine same-sample rule violations (via standard metrics that need the labels for evaluation purposes, like the mentioned AUROC).  We therefore
selected rules for which the corresponding datasets provide meaningful
semantic support rather than increasing syntactic depth for its own sake.
Scaling the implementation to a larger number of rules is a separate
computational question and can be studied independently from semantic
benchmark complexity.

\subsection{Baselines and Comparisons}
\label{sec:baselines}

We consider as baselines only \emph{rule-aware} detectors (use the same concept vocabulary and rule set as our method), since \emph{perception-only} anomaly detectors ignore rules. Unless otherwise stated, all baselines use the same backbone/encoder as the leaf concept bank for a controlled comparison.




\subsubsection{Independent-events probabilistic evaluator (\textsc{IndepProb}) (Rule-aware symbolic baseline).} Let $A,B$ denote leaf events (concepts) with predicted probabilities $p_A, p_B$ from the leaf bank. Let $\varphi$ be a rule formula built from leaves using $\neg,\wedge,\vee,\Rightarrow,\Leftrightarrow$. Assume leaf events are \textbf{independent} \emph{given} $x$, then one computes a soft satisfaction probability $P(\varphi)$ by recursion using the independent events prescriptions in App. \ref{indev}. For a general formula $\varphi$, evaluate bottom-up on the compiled rule DAG using these local identities (and applying edge-level negation by $p\mapsto 1-p$). The anomaly score is then $s_r(x)=1-P(\varphi_r\mid x)$.

\subsubsection{Semantic-loss (SEM) training baseline (semantic consistency from ordinary data).}
The core idea of SEM is to train with semantic structure as supervision \citep{Xu2018SemanticLoss}, rather than relying only on independent per-concept losses or on explicit real or synthetic anomalies. In our setting, this means encouraging the evaluator to learn rule-consistent compositions from ordinary samples: the model is exposed only to concept tuples that arise naturally in the data, and it must infer semantic compatibility or incompatibility from those same-image assignments. Thus, SEM captures the idea that logical or semantic structure can itself serve as a training signal, without requiring the explicit counterfactual/chimeric constructions that are central to our method.

\paragraph{Our implementation of SEM.}
We instantiate SEM in the variant that, in our view, provides the cleanest and fairest comparison with our full model while preserving the core semantic-training idea above. Concretely, SEM uses the same overall compositional evaluator pipeline as our method, but removes chimera-based supervision: all node concepts in a formula are instantiated from features extracted from the same normal image, and no real or synthesized anomalous or counterfactual compositions are introduced during training (since the application here is OOD, real anomalous samples are removed from the training). This keeps the rule compilation, evaluator structure, feature extractor, and test-time scoring protocol aligned with our full system, so that the main difference lies in the training signal itself. Consequently, the gap between SEM and the full model should be interpreted as quantifying the benefit of chimera-based compositional supervision beyond what can be learned from same-image semantic consistency alone.

\subsubsection{Monolithic root-only baseline.}
To isolate the contribution of our level-wise, node-local modular training scheme, we also consider a monolithic root-only baseline. In this variant, we remove all internal subtree gates and bottom-up supervision, and instead place a single MLP on top of the frozen leaf-bank outputs to predict the root truth value directly for each rule. The baseline therefore retains the same leaf encoder and rule-level target, but discards the explicit decomposition into local operators. We evaluate two versions: one trained only on normal same-image samples, and one additionally trained with chimera-based counterfactual compositions. The latter is the more informative ablation, since it tests whether chimera supervision alone is sufficient, or whether the full benefit of our method also depends on the level-wise and node-local modular structure.

\section{Results}

\begin{figure}
    \centering
    \includegraphics[width=1\linewidth]{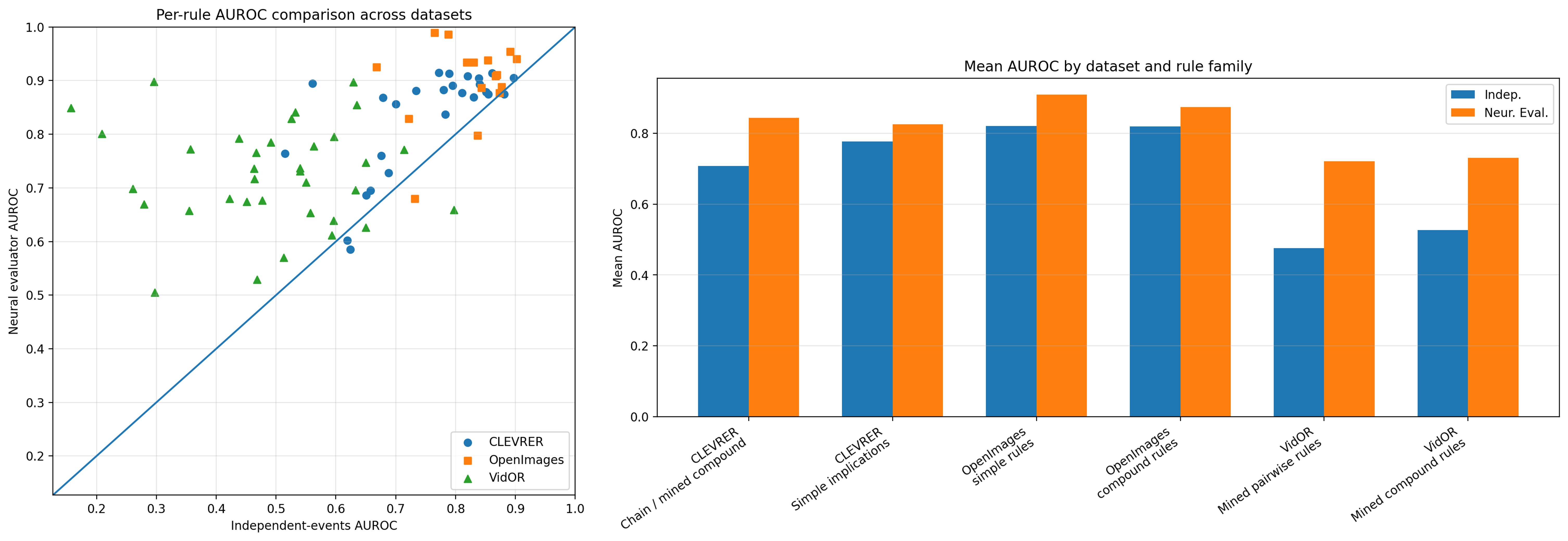}
    \caption{Comparison across datasets (left), and by dataset and rule family (right).}
    \label{fig:placeholder}
\end{figure}

\begin{table}[t]
\centering
\small
\setlength{\tabcolsep}{6pt}
\begin{tabular}{lccc}
\toprule
Dataset & Macro ROC-AUC & Macro AP & Macro Acc. \\
\midrule
CLEVRER    & 0.921 & 0.682 & 0.867 \\
OpenImages & 0.948 & 0.596 & 0.904 \\
VidOR      & 0.679 & 0.057 & 0.976 \\
\bottomrule
\end{tabular}
\caption{Leaf-bank performance on the evaluation split. For VidOR, macro AP is the more informative metric because many relation concepts are extremely sparse, so accuracy is inflated by class imbalance.}
\label{tab:leaf_quality}
\end{table}

\begin{table*}[t]
\centering
\small
\setlength{\tabcolsep}{4pt}
\begin{tabular}{llccccccc}
\toprule
Dataset & Rule family & \#Rules & Indep. & SEM & Mono-N & Mono-C & Neur. Eval. & Wins \\
\midrule
CLEVRER
& Simple implications    & 16 & 0.777 & 0.500 & 0.500 & 0.832 & 0.826 & 13/16 \\
& Chain \& Compound & 10 & 0.707 & 0.500 & 0.500 & 0.827 & 0.844 & 10/10 \\
& All rules    & 26 & 0.750 & 0.500 & 0.500 & 0.829 & 0.833 & 23/26 \\
Std*
& -    & - & 0.011 & 0.001 & 0.001 & 0.010 & 0.012 & - \\
\midrule
OpenImages
& Simple implications            & 11 & 0.821 & 0.500 & 0.500 & 0.906 & 0.910 & 10/11 \\
& Compound rules           & 5  & 0.820 & 0.500 & 0.500 & 0.880 & 0.874 & 4/5 \\
& All rules & 16 & 0.821 & 0.500 & 0.500 & 0.893 & 0.899 & 14/16 \\
Std*
& -    & - & 0.013 & 0.001 & 0.001 & 0.012 & 0.011 & - \\
\midrule
VidOR
& Simple implications  & 25 & 0.476 & 0.500 & 0.500 & 0.702 & 0.722 & 23/25 \\
& Compound rules  & 10 & 0.527 & 0.500 & 0.500 & 0.715 & 0.731 & 10/10 \\
& All rules             & 35 & 0.490 & 0.500 & 0.500 & 0.708 & 0.724 & 33/35 \\
Std*
& -    & - & 0.013 & 0.001 & 0.001 & 0.012 & 0.013 & - \\
\bottomrule
\end{tabular}
\caption{Rule-level anomaly AUROC (*in `All rules' on 3 runs). Indep.\ denotes the independent-events probabilistic evaluator; SEM is the same-image semantic-training ablation; Mono-N and Mono-C denote the monolithic root-only baseline trained, respectively, on normal samples only and with chimera-based counterfactual supervision. ``Wins'' counts how many rules are improved by the full neural evaluator relative to the independent-events baseline. For OpenImages, one degenerate rule with undefined AUROC was excluded from the aggregate.}
\label{tab:rule_aggregate}
\end{table*}

\begin{table}[t]
\centering
\scriptsize
\setlength{\tabcolsep}{4pt}
\begin{tabular}{llcc}
\toprule
Dataset & Rule & Indep. & Neur. Eval. \\
\midrule
CLEVRER & Chain: $\texttt{collide(brown, sphere)}\!\to\!\texttt{before\_half}\!\to\!\texttt{entered\_then\_collided}$  & 0.561 & 0.895 \\
CLEVRER & OR type: $\texttt{collide(cube,cyan)} \to (\texttt{before\_half} \vee \texttt{entered\_then\_collided})$ & 0.679 & 0.868 \\
OpenImages & $\texttt{Tableware} \to \texttt{Bottle}$ & 0.891 & 0.954 \\
OpenImages & $\texttt{Land\ vehicle} \to (\texttt{Bicycle} \wedge \texttt{Car})$ & 0.765 & 0.989 \\
VidOR & $\texttt{adult\!-\!in\_front\_of\!-\!baby} \to \texttt{baby\!-\!in\_front\_of\!-\!adult}$ & 0.467 & 0.766 \\
VidOR & $\texttt{obj:toy} \to (\texttt{child\!-\!next\_to\!-\!toy} \wedge \texttt{toy\!-\!in\_front\_of\!-\!child})$ & 0.526 & 0.829 \\
\bottomrule
\end{tabular}
\caption{Representative rule-level improvements. Full per-rule results are deferred to appendix \ref{detailtables}.}
\label{tab:qual_rules}
\end{table}

Table~\ref{tab:leaf_quality} first shows that the leaf concept bank is already reasonably strong on CLEVRER and OpenImages, whereas VidOR is substantially more challenging, especially for sparse relation concepts. Despite this, the proposed neural evaluator consistently improves rule-level anomaly AUROC over the independent-events baseline (Table~\ref{tab:rule_aggregate}). On the currently exported rules, the gain is +0.083 on CLEVRER (0.833 vs.\ 0.750, 23/26 wins), +0.078 on OpenImages (0.899 vs.\ 0.821, 14/16 wins after excluding one degenerate rule), and +0.234 on VidOR (0.724 vs.\ 0.490, 33/35 wins). The gains are especially pronounced on compositional rule families, including chain and mined compound rules on CLEVRER (+0.137) and both pairwise and compound mined rules on VidOR (+0.246 and +0.204, respectively). This is consistent with our hypothesis that feature-aware local composition captures dependencies that fixed probabilistic evaluation misses.

The SEM ablation collapses in the actual complex anomaly-detection benchmarks (this is quite less trivial than it may look like at first sight, cf. with the simpler MNIST scenario of Fig. \ref{fig:mnist}, where it \textit{does} actually retain some signal, likely thanks to the intrinsic uncertainty from the data). In this variant, we keep the same evaluator pipeline but train only on same-image normal samples and remove all chimera-based counterfactual compositions. Empirically, this drives the model to the trivial all-normal solution: it assigns near-zero anomaly score to essentially every sample, yielding AUROC $\approx 0.5$ across rules on the more complex datasets. This is consistent with our motivating hypothesis that, under realistic concept sparsity, ordinary normal data do not provide sufficient coverage of informative violating truth configurations; explicit chimera-based supervision is needed to learn a nontrivial anomaly signal.

\paragraph{Ablation: approximate independence vs.\ event entanglement.}
To probe whether the main failure mode of the fixed symbolic baseline is specifically its leaf-event independence assumption, we compare two closely related regimes. In \textbf{static CLEVR}, the leaves are mostly unary attribute-existence predicates (e.g., \texttt{blue\_sphere}, \texttt{green\_cube}, \texttt{metal\_any}), for which approximate factorization is often a reasonable first-order model. In \textbf{event-rich CLEVRER}, by contrast, we retain the same basic shape/color-style vocabulary but add temporally and relationally structured event predicates such as \texttt{collide(A,B)}, \texttt{collide\_before\_half(A,B)}, and \texttt{entered\_then\_collided(A,B)}. These predicates are strongly entangled by construction: collisions couple object pairs, temporal predicates depend on collision times, and several events co-occur non-independently.

This distinction is reflected clearly in the behavior of the independent-events evaluator. On static CLEVR, IndepProb is already essentially perfect on the tested rules (mean AUROC $=0.9995$; all rules in the range $0.999$--$1.000$). In contrast, on the currently exported event-rich CLEVRER rules, the same evaluator drops to mean AUROC $=0.750$, whereas the learned neural evaluator reaches $0.833$ on the same rule set. Thus, the degradation is not evidence that explicit symbolic structure is itself inadequate; rather, it strongly supports the view that the main weakness of the fixed symbolic baseline is \emph{misspecification of leaf-event independence}. Once collision-driven event structure is introduced, the factorized evaluator becomes structurally mismatched to the data, while the feature-aware local evaluator remains effective.


\begin{table}[t]
\centering
\small
\setlength{\tabcolsep}{6pt}
\begin{tabular}{lccc}
\toprule
Setup & Rule regime & IndepProb & Neur. Eval. \\
\midrule
CLEVR & static attribute rules & 0.9995 & --- \\
CLEVRER & event-rich rules & 0.750 & 0.833 \\
\bottomrule
\end{tabular}
\caption{Ablation contrasting a regime where approximate leaf independence is plausible (static CLEVR) with one where collision- and time-derived event predicates induce strong dependence (event-rich CLEVRER). On static CLEVR, the independent-events evaluator is already essentially perfect; on CLEVRER, its performance degrades substantially, while the learned evaluator remains stronger.}
\label{tab:independence_ablation}
\end{table}

\paragraph{Ablation of chimera training.} The SEM baseline is not only a baseline against competing approaches, but also an internal ablation of our own training scheme. In this variant, we keep the same overall compositional evaluator pipeline, but remove chimera-based supervision: all node concepts are instantiated using features drawn from the same image, and training uses only normal images, without the synthesized anomalous/counterfactual compositions introduced by chimera sampling. Thus, the comparison between the full model and SEM isolates the contribution of chimera training in a broad sense. The resulting performance gap indicates that the gain of the full method does not come merely from the architecture itself, but from the additional compositional supervision provided by chimera examples, which teaches the evaluator to detect semantically inconsistent concept combinations that are rarely or never observed in same-image normal training data alone.


\paragraph{Ablation of node-local, level-wise training.} The chimera-trained monolithic baseline performs similarly to the full neural evaluator. This suggests that, in the present regime, the dominant source of improvement is the chimera-based counterfactual supervision rather than the level-wise modular training scheme itself. In other words, once the model is exposed to informative synthetic rule violations, even a single root-level predictor can recover most of the anomaly signal. The result does not make the modular formulation unimportant---it still provides a more interpretable and reusable decomposition of rule evaluation, \textit{crucially without significant performance loss w.r.t. the monolithic case}---but it indicates that the principal performance bottleneck here was the lack of informative violating configurations in the training data.


\section{Discussion and Conclusion}
\label{sec:discussion}

We presented a method to learn and apply logical rules based on attributes that can be learned from the data. The key novelty is that complex rules can be learned without requiring all the logical combinations of attributes to be present in the data. We showed practical applications for anomaly detection in images and complex causality-constrained ones from video. 

The approach relies on two assumptions that may be violated in some regimes:
\begin{itemize}
  \item \textbf{Concept interface adequacy.} If the concept vocabulary is too weak (missing predicates needed to express true constraints), then rules become either vacuous or systematically violated, and anomaly detection degenerates.
  \item \textbf{Rule quality.} Mined rules can encode spurious correlations, especially in biased datasets. This can inflate anomaly scores for legitimate but underrepresented cases. Antecedent gating helps for implications, but it is not a complete fix for rule bias.
\end{itemize}

Our central premise is that many practical ``anomalies'' are not merely rare inputs, but \emph{constraint violations}. By compiling constraints into explicit DAGs and producing per-rule satisfaction probabilities via our chimera-trained evaluator, the detector returns not only a scalar score but also a structured explanation: which rules (and even which internal subclauses) are most violated. This is qualitatively different from perception-only OOD scores that are often hard to interpret, and it is difficult to obtain with monolithic end-to-end models without imposing explicit structure.

\bibliography{refs}
\bibliographystyle{tmlr}

\clearpage
\newpage
\appendix
\section*{Appendix}

\section{Method (in extended detail) and Algorithms}
\label{sec:methodapp}

\subsection{Problem setup and notation}
We assume a dataset of inputs and \emph{concept} annotations
\begin{equation}
\mathcal{D}=\{(x_i, y_i)\}_{i=1}^M,\qquad y_i\in\{0,1\}^N,
\end{equation}
where $y_{i,c}=1$ indicates that concept $c\in\{1,\dots,N\}$ is present in $x_i$ (multi-label).
We are also given a set of $R$ \emph{rules} $\{\mathcal{R}_r\}_{r=1}^R$. Each rule is compiled into a directed acyclic graph (DAG)
\begin{equation}
G_r=(V_r, E_r),
\end{equation}
whose leaves correspond to concept IDs and whose internal nodes are logical operators (IFF / IMPLIES / AND / OR). Edges may carry a negation flag, implementing literal negation at the child-to-parent interface.

Our goal is anomaly detection with \emph{semantic attribution}: for a test input $x$, output (i) an anomaly score $s(x)\in[0,1]$ and (ii) a decomposition into per-rule violation scores $\{s_r(x)\}$ (and optionally per-node scores).

\subsection{Leaf concept bank}
We first train a leaf concept bank that exposes:
\begin{align}
z &= E_\phi(x)\in\mathbb{R}^F,\\
\ell(x) &= (\ell_1(x),\dots,\ell_N(x))\in\mathbb{R}^N,\\
p(x) &= \sigma(\ell(x))\in(0,1)^N,
\end{align}
where $E_\phi$ is a shared encoder and $\ell_c(x)$ is a concept-specific logit head. Optionally, we apply post-hoc temperature scaling on logits (one scalar $T>0$) before the sigmoid.

In our implementation, the encoder feature $z$ is the basic evidence carrier for downstream rule evaluation. Concept probabilities $p(x)$ are additionally used for \emph{antecedent gating} in implication-style anomaly scoring (\S\ref{sec:anomaly_scoring}).

\subsection{Rule compilation as a DAG}
Each rule $\mathcal{R}_r$ is converted to a DAG $G_r$ with:
\begin{itemize}
  \item node attributes: $\texttt{mask}\in\{0,1\}$ (leaf vs internal), $\texttt{x}\in\{0,1,\dots,N\}$ (concept id for leaves), and $\texttt{y}\in\{1,2,3,4\}$ (operator code at internal nodes: IFF, IMPLIES, AND, OR),
  \item edge attributes: $\texttt{neg}\in\{-1,+1\}$ (negation) and (optionally) $\texttt{pos}\in\{0,1\}$ to enforce operand order for IMPLIES.
\end{itemize}
Commutative operators (IFF/AND/OR) treat children as an unordered multiset (canonicalized by sorting); IMPLIES is ordered (canonicalized by \texttt{pos} if present).

\paragraph{Hard semantics (for supervision).}
Given a concept label vector $y\in\{0,1\}^N$, we compute node-level ground-truth truth values $t_v(y)\in\{0,1\}$ by bottom-up propagation on $G_r$, applying edge negations at the child interface and then applying the exact Boolean operator at the parent. These $t_v(y)$ provide training targets for every internal node, not only the root.

\subsection{Subtree gates: learned, feature-aware logical composition}
\label{sec:subtree_gates}
Each internal node $v$ with arity $a_v$ is assigned a lightweight \emph{subtree gate} $g_{\theta_v}$ that maps child features and edge-negation flags to:
(i) a parent feature $h_v\in\mathbb{R}^F$ and
(ii) a satisfaction probability $\hat{t}_v\in(0,1)$.
Concretely, for children $c_1,\dots,c_{a_v}$ with features $h_{c_j}\in\mathbb{R}^F$ and negation flags $b_{j}\in\{0,1\}$,
\begin{align}
u_v &= \big[h_{c_1}\,\|\,b_1\,\|\,\cdots\,\|\,h_{c_{a_v}}\,\|\,b_{a_v}\big]\in\mathbb{R}^{a_v(F+1)},\\
h_v &= f_{\theta_v}(u_v)\in\mathbb{R}^{F},\\
\hat{t}_v &= \sigma(w_v^\top h_v + \beta_v)\in(0,1).
\end{align}
Intuitively, $g_{\theta_v}$ is a learned operator specialized to the local connective at $v$, but it operates in a feature space that can represent richer evidence than a scalar leaf probability.

\paragraph{Leaf features.}
For a given $x$, we initialize each leaf node $v$ that corresponds to concept $c(v)$ with a feature derived from the leaf-bank encoder:
\begin{equation}
h_v \leftarrow z = E_\phi(x).
\end{equation}
Although this means all leaves share the same base feature vector, the \emph{rule structure and gate identities} specify which leaf positions correspond to which concepts; gates learn to extract concept-relevant evidence from $z$ in a rule-conditional way. (This choice makes the evaluator robust to noisy scalar concept probabilities, while retaining concept semantics through the graph wiring and supervision.)

\subsection{Bottom-up, level-wise training with internal-node supervision}
\label{sec:levelwise_training}
Directly training all gates jointly can be unstable because higher nodes depend on learned representations of lower subtrees. We therefore train \emph{bottom-up} by depth.

Let $\mathrm{depth}(v)$ be the longest path length from any leaf to $v$, and let $\mathcal{V}_d=\{v:\mathrm{depth}(v)=d\}$. We train levels in increasing depth $d=1,2,\dots$ (see Algorithm \ref{alg:training}):
\begin{enumerate}
\item For each mini-batch $\{(x_i,y_i)\}_{i=1}^B$, compute hard truth targets $t_v(y_i)$ for all nodes by exact propagation (Algorithm \ref{alg:propagate-hard-truths-core}). This is where the choice of representing rules by graphs becomes particularly useful and elegant at the implementation level: we use DGL's $\texttt{dgl.topological\_nodes\_generator}$ to generate node frontiers using topological traversal (each item is a tensor recording the nodes from bottom level to the roots).
\item Compute leaf encoder features $z_i=E_\phi(x_i)$ and initialize leaf node features.
\item Propagate through already-trained lower-depth gates (and keep them fixed) to obtain child features for nodes in $\mathcal{V}_d$.
\item For each $v\in\mathcal{V}_d$, update gate parameters by minimizing node-wise BCE:
\begin{equation}
\mathcal{L}_v = \frac{1}{B}\sum_{i=1}^B \mathrm{BCE}\big(\hat{t}_v(x_i),\ t_v(y_i)\big).
\end{equation}
\end{enumerate}
This \emph{internal supervision} is crucial: the model learns to implement logical composition locally, rather than only learning a monolithic ``rule classifier'' at the root.

\subsection{Chimera negative training: enforcing compositionality and preventing shortcut learning}
\label{sec:chimera}
A key failure mode is \emph{shortcut learning}: a gate (or an entire rule) can be predicted directly from global visual features, without respecting the intended operator semantics. We introduce \emph{chimera negatives} to force gates to behave compositionally.

Consider a binary node $v$ with (ordered) children $(\ell, r)$ and operator $\mathrm{op}(v)$. For a batch of size $B$, choose a permutation $\pi$ with $\pi(i)\neq i$ (e.g., a random nontrivial cyclic shift). We form \emph{chimera pairs} by combining the left child from sample $i$ with the right child from a different sample $\pi(i)$:
\begin{equation}
(u_v^{\text{chim}})_i := \big[h_\ell(x_i)\,\|\,b_\ell\,\|\,h_r(x_{\pi(i)})\,\|\,b_r\big].
\end{equation}
Targets are computed \emph{semantically} from the corresponding hard truths (including edge negations):
\begin{equation}
(t_v^{\text{chim}})_i := \mathrm{op}(v)\big(t_\ell(y_i),\ t_r(y_{\pi(i)})\big)\in\{0,1\}.
\end{equation}
Training batches can include: (i) same-image pairs, optionally restricted to samples where the full rule holds (``AD-strict''), and (ii) chimera pairs (Algorithm \ref{alg:training}). This construction breaks correlations that enable shortcut solutions, because the two operands are decoupled across samples while the supervision remains the exact logical composition. In practice, chimera training substantially improves transfer of learned subtrees across rules and reduces overfitting to rule-specific visual templates: only the true relevant features for the rule to hold or not end being recognized.

\paragraph{Novel insight.}

Chimera training moves counterfactual mixing to the operand level, supplying informative truth assignments even when specific rule outcomes have little or no support in the observed training data.

\subsection{Lineage-aware caching: safe reuse of learned subtrees}\label{sec}To scale to many rules, we reuse trained gates across rule graphs whenever the corresponding subtrees match. However, naive caching is unsafe: if the upstream encoder changes, the meaning of features changes. We therefore define a \emph{lineage-aware} cache key $K(v)$ recursively:

\begin{align}
K(v) &=
\begin{cases}
\texttt{LEAF}(c(v)\,|\,\texttt{enc}=\mathrm{fp}(E_\phi)) & \text{if $v$ is a leaf,}\\
\texttt{OP}\_{\mathrm{op}(v)}\big(\widetilde{K}(c_1),\dots,\widetilde{K}(c_{a_v})\big)\,|\,\texttt{arch}\,|\,F
& \text{if $v$ is internal,}
\end{cases}\\
\widetilde{K}(c_j) &= 
\begin{cases}
K(c_j) & \text{if edge $(c_j\!\to\! v)$ is non-negated,}\\
\texttt{!}K(c_j) & \text{if edge $(c_j\!\to\! v)$ is negated,}
\end{cases}
\end{align}
where $\mathrm{fp}(E_\phi)$ is a short fingerprint (hash) of encoder weights, $\texttt{arch}$ tags the gate architecture version, and $F$ is the feature dimension. We hash $K(v)$ to obtain a stable filename and store the gate state dict. A cached gate is reused iff the key matches exactly.

\subsection{Algorithms}

\begin{algorithm}[!h]
\caption{$\mathrm{HardOp}(op,\textsf{vals})$ --- exact Boolean semantics used by \textsc{PropagateHardTruths}}
\label{alg:propagate-hard-truths-hardop}
\begin{algorithmic}[1]
\REQUIRE Operator code $op\in\{1,2,3,4\}$ and child truth list $\textsf{vals}[1..m]$ with $m=|\textsf{vals}|$.
\ENSURE Hard truth $t\in\{0,1\}$.

\IF{$op=1$} \STATE \textbf{(IFF)}
  \STATE $t\leftarrow 1$ \hfill (empty list returns 1)
  \FOR{$j=2$ to $m$}
    \IF{$\textsf{vals}[j]\neq \textsf{vals}[1]$}
      \STATE $t\leftarrow 0$
    \ENDIF
  \ENDFOR
\ELSE
  \IF{$op=2$} \STATE \textbf{(IMPLIES)}
    \IF{$m\neq 2$}
      \STATE $t\leftarrow 1$
    \ELSE
      \STATE $a\leftarrow \textsf{vals}[1]$;\ \ $b\leftarrow \textsf{vals}[2]$
      \STATE $t\leftarrow (1-a)\ \textbf{or}\ b$
    \ENDIF
  \ELSE
    \IF{$op=3$} \STATE \textbf{(AND)}
      \STATE $t\leftarrow 1$ \hfill (empty list returns 1)
      \FOR{$j=1$ to $m$}
        \IF{$\textsf{vals}[j]=0$}
          \STATE $t\leftarrow 0$
        \ENDIF
      \ENDFOR
    \ELSE
      \IF{$op=4$} \STATE \textbf{(OR)}
        \STATE $t\leftarrow 0$ \hfill (empty list returns 0)
        \FOR{$j=1$ to $m$}
          \IF{$\textsf{vals}[j]=1$}
            \STATE $t\leftarrow 1$
          \ENDIF
        \ENDFOR
      \ELSE
        \STATE $t\leftarrow 0$
      \ENDIF
    \ENDIF
  \ENDIF
\ENDIF

\RETURN $t$.
\end{algorithmic}
\end{algorithm}

\begin{algorithm}[!h]
\caption{\textsc{PropagateHardTruths}$(G_r, y_i)$ --- hard-truth propagation (core loop)}
\label{alg:propagate-hard-truths-core}
\begin{algorithmic}[1]
\REQUIRE Rule DAG $G_r=(V,E)$ with node attrs:
$\texttt{mask}[v]\in\{0,1\}$ (1=leaf),
$\texttt{x}[v]\in\{0,1,\dots,K\}$ (concept id),
$\texttt{y}[v]\in\{1,2,3,4\}$ (1=IFF, 2=IMPLIES, 3=AND, 4=OR);
and edge attr $\texttt{neg}[e]\in\{-1,+1\}$ (optional).
\REQUIRE Concept hard labels $y_i\in\{0,1\}^{K}$ indexed by concept id $1..K$.
\ENSURE Node hard truth vector $t\in\{0,1\}^{|V|}$, stored as $G_r.\texttt{ndata['truth\_value']}$.

\STATE Initialize $t[v]\leftarrow 0$ for all $v\in V$.
\STATE \textbf{Leaf init by concept id}
\FOR{each node $v\in V$}
  \IF{$\texttt{mask}[v]=1$}
    \STATE $c\leftarrow \texttt{x}[v]$
    \IF{$c>0$}
      \STATE $t[v]\leftarrow y_i[c]$
    \ENDIF
  \ENDIF
\ENDFOR

\STATE \textbf{Bottom-up pass in topological order}
\STATE Let $\pi$ be a topological ordering of $V$ where children precede parents.
\FOR{each node $v$ in $\pi$}
  \IF{$\texttt{mask}[v]=0$}
    \STATE $(\texttt{src},\texttt{eid})\leftarrow G_r.\texttt{in\_edges}(v,\texttt{form='all'})$.
    \STATE $m \leftarrow |\texttt{src}|$;\ \ create array $\textsf{vals}[1..m]$.
    \FOR{$j=1$ to $m$}
      \STATE $u \leftarrow \texttt{src}[j]$;\ \ $e\leftarrow \texttt{eid}[j]$;\ \ $\textsf{vals}[j]\leftarrow t[u]$.
      \IF{$\texttt{neg}$ exists \AND $\texttt{neg}[e]=-1$}
        \STATE $\textsf{vals}[j]\leftarrow 1-\textsf{vals}[j]$.
      \ENDIF
    \ENDFOR
    \STATE $op \leftarrow \texttt{y}[v]$.
    \STATE $t[v] \leftarrow \mathrm{HardOp}(op,\textsf{vals})$ \hfill (Alg.~\ref{alg:propagate-hard-truths-hardop})
  \ENDIF
\ENDFOR

\STATE $G_r.\texttt{ndata['truth\_value']}\leftarrow t$.
\RETURN $t$.
\end{algorithmic}
\end{algorithm}

\vspace{4pt}

\begin{algorithm}[!h]
\caption{Training: Leaf Bank + Cached Subtree Gates (Level-wise with Chimera Negatives)}
\label{alg:training}
\begin{algorithmic}[1]
\REQUIRE Training set $\mathcal{D}_{\text{train}}=\{(x_i,y_i)\}$ with concept labels $y_i\in\{0,1\}^K$; rule set $\{\mathcal{R}_r\}_{r=1}^R$ compiled into DAGs $\{G_r\}$; feature dim $F$; gate architecture tag \texttt{arch}; negatives mode \texttt{neg\_mode}; cache directory $\mathcal{C}$
\ENSURE Leaf concept bank parameters $\phi$; cached gate parameters in $\mathcal{C}$

\vspace{2pt}
\STATE \textbf{(A) Train leaf concept bank}
\STATE Initialize encoder+heads $\phi$ (shared encoder $E_\phi$, $K$ sigmoid heads)
\FOR{epoch $=1$ to $E_{\text{leaf}}$}
  \FOR{mini-batch $\{(x_i,y_i)\}_{i=1}^B$}
    \STATE $z_i \leftarrow E_\phi(x_i)$; logits $\ell_i\leftarrow H_\phi(z_i)$
    \STATE Update $\phi$ by minimizing multi-label BCEWithLogits$(\ell_i,y_i)$ (optionally with \texttt{pos\_weight})
  \ENDFOR
\ENDFOR
\IF{\texttt{use\_temp\_scaling}}
  \STATE Fit scalar temperature $T$ on held-out logits by minimizing BCE$(\sigma(\ell/T),y)$; store calibrator
\ENDIF

\vspace{2pt}
\STATE \textbf{(B) Train rule evaluators as reusable subtree gates}
\STATE Compute encoder fingerprint $\mathrm{fp}\leftarrow \mathrm{Fingerprint}(E_\phi)$
\FOR{each rule graph $G_r$}
  \STATE Let $D_{\max}\leftarrow$ maximum internal-node depth in $G_r$
  \FOR{depth $d=1$ to $D_{\max}$}
    \STATE Let $\mathcal{V}_d\leftarrow\{v\in V_r:\mathrm{depth}(v)=d\}$
    \STATE Load or initialize gates $\{g_{\theta_v}\}_{v\in \mathcal{V}_d}$ from cache using lineage key $K(v;\mathrm{fp},\texttt{arch},F)$
    \FOR{mini-batch $\{(x_i,y_i)\}_{i=1}^B$}
      \STATE \textbf{Hard node targets:} $t_{v,i}\leftarrow \mathrm{PropagateHardTruths}(G_r, y_i)\ \forall v$
      \STATE \textbf{Leaf evidence:} $z_i\leftarrow E_\phi(x_i)$; $p_i\leftarrow \sigma(\ell_i)$ from leaf heads
      \STATE Initialize leaf node features $h_{\text{leaf}} \leftarrow z_i$ (placed by concept id); leaf probs $\hat t_{\text{leaf}}\leftarrow p_i$
      \STATE Propagate already-trained lower-depth gates to compute $(h_{c,i},\hat t_{c,i})$ for all children of nodes in $\mathcal{V}_d$
      \FOR{each node $v\in\mathcal{V}_d$}
        \STATE Build \textbf{same-image} inputs $u^{\text{same}}_{v,i}=\big[h_{c_1,i}\|b_1\|\cdots\|h_{c_a,i}\|b_a\big]$
        \STATE Set targets $t^{\text{same}}_{v,i}=t_{v,i}$
        \IF{\texttt{neg\_mode} uses chimera \AND $v$ is binary}
          \STATE Choose a permutation $\pi$ with $\pi(i)\neq i$
          \STATE Build \textbf{chimera} inputs $u^{\text{chim}}_{v,i}=\big[h_{\ell,i}\|b_\ell\|h_{r,\pi(i)}\|b_r\big]$
          \STATE Set targets $t^{\text{chim}}_{v,i}=\mathrm{Op}_v\!\big(t_{\ell,i},t_{r,\pi(i)}\big)$ (with edge negations applied in the hard truths)
          \STATE Concatenate training pairs $(u_{v},t_v)\leftarrow (u^{\text{same}},t^{\text{same}})\cup(u^{\text{chim}},t^{\text{chim}})$ (optionally filter same-image pairs, e.g.\ ``AD-strict'')
        \ELSE
          \STATE $(u_{v},t_v)\leftarrow (u^{\text{same}},t^{\text{same}})$
        \ENDIF
        \STATE Update $\theta_v$ by minimizing $\frac{1}{|u_v|}\sum \mathrm{BCE}\big(g_{\theta_v}(u_v), t_v\big)$
      \ENDFOR
    \ENDFOR
    \STATE Save gates $\{g_{\theta_v}\}$ to cache $\mathcal{C}$ under lineage keys $K(v;\mathrm{fp},\texttt{arch},F)$
  \ENDFOR
\ENDFOR
\RETURN $\phi$ and cache $\mathcal{C}$
\end{algorithmic}
\end{algorithm}

\vspace{4pt}

\begin{algorithm}[!h]
\caption{\textsc{PredictRoot}$(G_r, z, p;\mathcal{C})$}
\label{alg:predict-root-fn}
\begin{algorithmic}[1]
\REQUIRE Rule DAG $G_r=(V,E)$ with node attrs $\texttt{mask},\texttt{x},\texttt{y}$ and edge attrs $\texttt{neg}$ (and optionally $\texttt{pos}$ for ordering IMPLIES children).
\REQUIRE Leaf evidence for one sample: feature vector(s) $z$ and leaf concept probabilities $p\in(0,1)^{K}$ from the leaf bank.
\REQUIRE Cache/registry $\mathcal{C}$ for subtree gates $g_v$ at internal nodes $v$ (load by lineage key in code).
\ENSURE Root probability $\hat t_{\text{root}}\in(0,1)$.

\STATE Initialize node feature tensor $h[v]\leftarrow 0$ and node prob $\hat t[v]\leftarrow 0$ for all $v\in V$.
\STATE \textbf{(Place leaf probabilities and leaf features by concept id)}
\FOR{each node $v\in V$}
  \IF{$\texttt{mask}[v]=1$}
    \STATE $c\leftarrow \texttt{x}[v]$.
    \IF{$c>0$}
      \STATE $\hat t[v]\leftarrow p_c$.
      \STATE $h[v]\leftarrow z_c$ \hfill (in code: take the feature slice associated with concept $c$)
    \ENDIF
  \ENDIF
\ENDFOR

\STATE \textbf{(Bottom-up gate propagation level-by-level)}
\STATE Let $D_{\max}$ be the maximum internal-node depth in $G_r$; let $\mathcal{V}_d$ be nodes at depth $d$ (as precomputed in the trainer).
\FOR{$d=1$ to $D_{\max}$}
  \FOR{each node $v\in \mathcal{V}_d$}
    \STATE Load gate $g_v$ for $v$ (runtime registry if present; else load from $\mathcal{C}$ by the node's cache key).
    \STATE Query incoming edges: $(\texttt{src},\texttt{eid})\leftarrow G_r.\texttt{in\_edges}(v,\texttt{form='all'})$.
    \STATE If $op=\texttt{y}[v]=2$ (IMPLIES), reorder the two incoming edges using $\texttt{pos}$ if available (antecedent first); otherwise keep the incoming-edge order.
    \STATE Build gate input by concatenating each child feature with a negation bit:
    \STATE \hspace{1em}Create empty vector $u_v$.
    \FOR{$j=1$ to $|\texttt{src}|$}
      \STATE $c\leftarrow \texttt{src}[j]$;\ \ $e\leftarrow \texttt{eid}[j]$.
      \IF{$\texttt{neg}$ exists \AND $\texttt{neg}[e]=-1$}
        \STATE $b\leftarrow 1$.
      \ELSE
        \STATE $b\leftarrow 0$.
      \ENDIF
      \STATE Append $\big[h[c]\ \|\ b\big]$ to $u_v$ \hfill (concatenate along feature dimension)
    \ENDFOR
    \STATE Forward gate: $(h[v],\hat t[v])\leftarrow g_v(u_v)$.
  \ENDFOR
\ENDFOR

\STATE Let $\text{root}$ be the unique node with out-degree $0$.
\RETURN $\hat t[\text{root}]$.
\end{algorithmic}
\end{algorithm}

\vspace{4pt}

\begin{algorithm}[!h]
\caption{Inference: Rule Satisfaction, Violation Attribution, and Anomaly Scoring}
\label{alg:inference}
\begin{algorithmic}[1]
\REQUIRE Test input $x$; trained leaf bank $\phi$ (and optional temperature $T$); rule graphs $\{G_r\}$; gate cache $\mathcal{C}$; aggregation mode \texttt{Agg}; implication gate threshold $\tau$
\ENSURE Anomaly score $s(x)$; per-rule scores $\{s_r(x)\}$; top-$k$ violated rules

\STATE Compute encoder feature $z\leftarrow E_\phi(x)$
\STATE Compute concept probabilities $p\leftarrow \sigma(\ell/T)$ (use $T{=}1$ if no calibration)
\FOR{each rule graph $G_r$}
  \STATE Load any missing gates from cache $\mathcal{C}$ (by lineage key) into runtime registry
  \STATE \textbf{Predict root satisfaction:} $p_r(x)\leftarrow \mathrm{PredictRoot}(G_r, z, p;\mathcal{C})$ \hfill// bottom-up gate propagation
  \IF{$G_r$ is an implication rule with antecedent concept $A_r$ (or antecedent subgraph)}
    \STATE $a_r \leftarrow p(A_r)$ \hfill// antecedent probability from leaf bank
    \STATE $g(a_r)\leftarrow \max(0,a_r-\tau)/(1-\tau)$ \hfill// $\tau=0$ gives identity
    \STATE \textbf{Antecedent-weighted violation:} $s_r(x)\leftarrow g(a_r)\cdot (1-p_r(x))$
  \ELSE
    \STATE \textbf{Violation:} $s_r(x)\leftarrow 1-p_r(x)$
  \ENDIF
\ENDFOR

\STATE \textbf{Aggregate:} $s(x)\leftarrow \texttt{Agg}(\{s_r(x)\}_{r=1}^R)$ \hfill// e.g.\ max/mean/geo/learned
\STATE \textbf{Attribution:} return top-$k$ rules by descending $s_r(x)$ (and optionally the most-violated internal nodes)
\RETURN $s(x)$ and $\{s_r(x)\}$
\end{algorithmic}
\end{algorithm}
\clearpage

\section{Datasets and concept vocabularies}
We evaluate on three vision benchmarks where (i) a multi-label concept inventory is available (or can be induced), and (ii) logical constraints over these concepts are meaningful.

\paragraph{CLEVR (images).}
We use CLEVR images and their ground-truth scene annotations \cite{Johnson2017CLEVR}. Each concept is a unary predicate of the form ``$\exists$ object with attributes matching a filter''. Concretely, our base concept bank contains $K{=}6$ attribute filters (e.g., \texttt{blue\_sphere}, \texttt{metal\_any}, \texttt{gray\_cyl}), and an image-level label $y_k\in\{0,1\}$ is positive iff at least one object in the scene satisfies the corresponding filter.

\paragraph{CLEVRER (videos).}
We use CLEVRER and its structured annotations \cite{Yi2020CLEVRER}. We reuse the same $K{=}6$ base attribute-filter concepts for object existence, computed from the per-video object annotations (positive iff the video contains at least one object matching the filter).
Optionally (and in the ``event-rich'' setting), we extend the concept vocabulary with event predicates constructed from CLEVRER event annotations: \texttt{enter(A)}, \texttt{exit(A)}, \texttt{collide(A,B)}, \texttt{collide\_before\_half(A,B)}, and \texttt{entered\_then\_collided(A,B)} where $A,B$ are attribute filters. We cap the number of attribute-pair instantiations (hyperparameter \texttt{max\_event\_pairs}) to keep the rule set tractable.

\paragraph{Open Images.}
We use Open Images V4 annotations \cite{Kuznetsova2018OpenImagesV4} and restrict to a manageable concept set by selecting the top-$K$ most frequent detection classes in the validation bounding-box CSV (default $K{=}50$).
A concept label is positive iff an image has \emph{at least one} bounding box of that class.
Since Open Images does not provide a canonical train/val split for this exact ``consistency anomaly'' task, we split the validation images into a \emph{rule/gate-train} partition and an \emph{eval} partition (default \texttt{val\_train\_frac}=0.9).

\paragraph{VidOR.}
We evaluate on the VidOR (Video Object Relation) dataset~\cite{shang2019annotating}, a large-scale collection of $10{,}000$ user-generated videos (98.6 hours) with spatio-temporal annotations of object trajectories and relation instances (80 object categories, 50 relation predicates), using the official split of 7,000/835/2,165 videos for train/val/test. 
Since test annotations are not fully available, we follow a train$\rightarrow$val protocol.
From training annotations, we construct a multi-label concept vocabulary of $K$ leaves consisting of (i) \texttt{obj:$c$} for the top-$K_{\text{obj}}$ most frequent object categories and (ii) \texttt{rel:$s$-$p$-$o$} for the top-$K_{\text{rel}}$ most frequent relation triplets (subject category $s$, predicate $p$, object category $o$), keeping only triplets with at least a minimum support count.
For each video, we uniformly sample $T$ frames and form a multi-hot label vector $y\in\{0,1\}^K$: an object leaf is positive if its annotated trajectory is present in any sampled frame; a relation leaf is positive if any annotated relation instance is active in at least one sampled frame (with temporal extent $[t_{\text{begin}},t_{\text{end}})$). 
Whenever a relation is labeled positive, we additionally mark its subject and object categories as present, enforcing the structural constraint \texttt{rel:$s$-$p$-$o$}$\Rightarrow$(\texttt{obj:$s$}$\wedge$\texttt{obj:$o$}) at the ground-truth label level.
We train a video leaf bank as a multi-label classifier with a shared 2D CNN applied per-frame and mean temporal pooling to obtain a clip embedding, followed by $K$ sigmoid heads; optional temperature scaling is fit on the training split.
Rules are mined from training labels (including depth-2 compounds) and compiled into logical DAGs; subtree gates are trained level-wise using internal-node Boolean supervision.
At evaluation time, we report (i) concept prediction metrics on the validation set and (ii) consistency anomaly detection, where a clip is labeled anomalous iff it violates at least one rule under hard Boolean evaluation of the ground-truth concept vector.

\subsection{Rule construction from annotations}
Across datasets, our constraints are expressed as small boolean formula graphs (simple implications and depth-2 compound rules), compiled into DGL graphs and evaluated both (i) \emph{hard} on ground-truth concept labels and (ii) \emph{soft} on predicted concept probabilities.

\paragraph{Handwritten seed and compound rules (CLEVRER).}
For CLEVRER we include a small set of \emph{seed} implications and a small set of \emph{compound} depth-2 formulas (conjunctions/disjunctions under an implication).
We additionally include event-structure constraints consistent with the event definitions, e.g.
\begin{align}
\texttt{collide(A,B)} &\Rightarrow \texttt{collide\_before\_half(A,B)}.
\end{align}

\paragraph{Mined pairwise implication rules.}
We mine high-confidence implications directly from the training concept labels by co-occurrence statistics.
For each ordered pair $(A,B)$ we estimate
\[
\widehat{P}(B{=}1\mid A{=}1)=\frac{\#(A\wedge B)}{\#(A)}.
\]
We keep candidates only if $A$ has sufficient support, $\#(A)/N \ge \texttt{support\_thresh}$ (default 0.05), and then add:
(i) an implication $A\Rightarrow B$ if $\widehat{P}(B\mid A)\ge \texttt{confidence\_pos}$ (default 0.995), or
(ii) an exclusion $A\Rightarrow \neg B$ if $\widehat{P}(B\mid A)\le \texttt{confidence\_neg}$ (default 0.005).
We cap the number of mined rules (default \texttt{max\_rules}=25).

\paragraph{Taxonomy- and part-based rules + mined pairs (Open Images).}
We build implication rules from the Open Images class hierarchy and (optionally) part-of relations, restricted to the top-$K$ selected classes.
Because closure assumptions can render some ``natural direction'' constraints tautological (depending on how labels are completed), we also evaluate an \emph{inverted} constraint direction that is explicitly nontrivial under label closure (e.g., coarse $\Rightarrow$ fine, whole $\Rightarrow$ part).
Additionally, we mine high-confidence co-occurrence rules from the bounding-box validation CSV with thresholds \texttt{min\_support} (default 200 images) and \texttt{min\_conf} (default 0.99), and we generate a capped number of sibling-based compound rules per parent (\texttt{per\_parent\_pair\_limit}).

\paragraph{Upward closure for Open Images supervision and its implications.}
Open Images detection annotations are not exhaustive across the taxonomy: an image may be annotated with a fine-grained class while omitting its ancestors. If missing ancestors were treated as negatives, training a multi-label concept bank would inject systematic false negatives for parent concepts (e.g., \texttt{Labrador}=1 but \texttt{Dog}=0), forcing the model to suppress parent predictions that are logically entailed by labeled descendants. To avoid this pathology, we apply an \emph{upward hierarchical closure} to the ground-truth labels before training: whenever a class is present, all its ancestors in the provided hierarchy are marked present as well. This makes supervision consistent with the intended semantics that ancestors represent coarse presence.

A direct consequence is that \emph{child$\rightarrow$parent implications become tautological in the closed label space} (there are no hard violations by construction), and therefore cannot define informative ``consistency anomalies.'' For Open Images we instead evaluate constraints that remain nontrivial under upward closure, such as \emph{inverted} implications (coarse$\rightarrow$fine and whole$\rightarrow$part) with antecedent gating, and mined high-confidence co-occurrence implications. In this setting, flagged violations should be interpreted as \emph{missing-detail / semantic inconsistency signals} relative to the learned concept bank and the chosen constraint family, rather than contradictions of the closed taxonomy itself.

\paragraph{Open-set mixture in evaluation and label-space caveats (Open Images).}
Although our Open Images concept inventory is restricted to a Top-$K$ subset (for tractable multi-label learning), the \emph{evaluation distribution} is not purely closed-set: the validation split naturally contains many instances of object subtypes outside the Top-$K$ leaf vocabulary (e.g., unmodeled vehicle subclasses). Consequently, our reported AUROCs are measured on a realistic \emph{mixture} of in-vocabulary and out-of-vocabulary content. This mixed setting introduces a protocol-specific label issue for implication-based constraints: when an image contains an out-of-Top-$K$ descendant of an antecedent parent, the antecedent may be absent from the Top-$K$ ground-truth vector used to compute rule truths (even after upward closure restricted to Top-$K$), rendering the implication vacuously true in the evaluation labels. This creates conservative label noise that can underestimate rule-violation detection performance. 

To mitigate this, we filter evaluation samples where the image contains an out-of-vocabulary descendant of a rule antecedent (in the full Open Images label space), but that antecedent is absent in the Top-$K$ ground-truth vector used for rule-truth computation. This situation arises because upward closure is applied only within the Top-$K$ label space: descendants outside Top-$K$ cannot trigger their Top-$K$ ancestors, so the antecedent is spuriously missing and the implication is labeled vacuously true. Removing these cases reduces label-noise that would otherwise penalize methods for correctly inferring the antecedent from visual evidence.

A separate caveat concerns \emph{hierarchical closure itself}: upward closure can activate coarse ancestors based on labeled descendants, and depending on the ontology semantics, this may not coincide with ``whole object visibly present'' in the scene (and may introduce its own inconsistencies). This is a generic consequence of adopting closure on a given hierarchy (shared by many closure-based pipelines) rather than a pathology specific to our Top-$K$ implication evaluation protocol; our evaluation correction targets the former protocol-induced vacuity/coverage issue, not the closure assumption.

\section{Detailed Results Tables}\label{detailtables}

\subsection{CLEVRER -- Detailed Results Tables}

\begingroup
\setlength{\tabcolsep}{3pt}
\renewcommand{\arraystretch}{1.05}
\tiny
\begin{longtable}{p{0.46\columnwidth}rrrr}
\caption{CLEVRER leaf-bank evaluation on the validation split. ``Prev.'' denotes the number of positive validation videos for the class.}
\label{tab:clevrer_leaf_perclass_onecol}\\
\toprule
Class & Prev. & ROC-AUC & AP & Acc.@0.5 \\
\midrule
\endfirsthead

\toprule
Class & Prev. & ROC-AUC & AP & Acc.@0.5 \\
\midrule
\endhead

\midrule
\multicolumn{5}{r}{Continued on next page}
\endfoot

\bottomrule
\endlastfoot

blue\_sphere & 2263 & 0.991 & 0.989 & 0.956 \\
red\_sphere & 2210 & 0.988 & 0.984 & 0.940 \\
green\_cube & 2077 & 0.948 & 0.893 & 0.900 \\
yellow\_cyl & 2212 & 0.963 & 0.929 & 0.918 \\
metal\_any & 4837 & 0.544 & 0.971 & 0.852 \\
gray\_cyl & 2178 & 0.961 & 0.925 & 0.864 \\
enter(shape=sphere) & 2817 & 0.818 & 0.832 & 0.694 \\
exit(shape=sphere) & 652 & 0.904 & 0.542 & 0.871 \\
\end{longtable}
\endgroup

\begingroup
\setlength{\tabcolsep}{3pt}
\renewcommand{\arraystretch}{1.08}
\tiny
\begin{longtable}{p{0.06\columnwidth}p{0.66\columnwidth}r}
\caption{Per-rule CLEVRER results for the \textbf{Independent-Events Probabilistic evaluator}. For readability, repeated predicate arguments within the same row are shown only on first occurrence.}
\label{tab:clevrer_indep_perrule_onecol}\\
\toprule
ID & Rule & AUROC \\
\midrule
\endfirsthead

\toprule
ID & Rule & AUROC \\
\midrule
\endhead

\midrule
\multicolumn{3}{r}{Continued on next page}
\endfoot

\bottomrule
\endlastfoot

01 &
\makecell[tl]{[Chain] $(\text{collide}(\text{shape=sphere},\text{color=brown}) \to \text{collide\_before\_half})$\\
$\wedge\;(\text{collide\_before\_half} \to \text{entered\_then\_collided})$}
& 0.561 \\

02 &
\makecell[tl]{[Chain] $(\text{collide}(\text{shape=cube},\text{color=cyan}) \to \text{collide\_before\_half})$\\
$\wedge\;(\text{collide\_before\_half} \to \text{entered\_then\_collided})$}
& 0.795 \\

03 &
\makecell[tl]{[Chain] $(\text{collide}(\text{shape=sphere},\text{color=green}) \to \text{entered\_then\_collided})$\\
$\wedge\;(\text{entered\_then\_collided} \to \text{collide\_before\_half})$}
& 0.780 \\

04 &
\makecell[tl]{[Chain] $(\text{collide}(\text{shape=cube},\text{color=blue}) \to \text{entered\_then\_collided})$\\
$\wedge\;(\text{entered\_then\_collided} \to \text{collide\_before\_half})$}
& 0.840 \\

05 &
\makecell[tl]{$\text{collide}(\text{shape=sphere},\text{shape=cube}) \to (\text{collide\_before\_half} \wedge \text{entered\_then\_collided})$\\
{[mined-AND; conf=0.51, hits=1041, viol=1015]}}
& 0.676 \\

06 &
\makecell[tl]{$\text{collide}(\text{shape=sphere},\text{shape=cylinder}) \to (\text{collide\_before\_half} \wedge \text{entered\_then\_collided})$\\
{[mined-AND; conf=0.49, hits=1148, viol=1212]}}
& 0.688 \\

07 &
\makecell[tl]{$\text{collide}(\text{shape=cube},\text{color=cyan}) \to (\text{collide\_before\_half} \vee \text{entered\_then\_collided})$\\
{[mined-OR; conf=0.95, hits=726, viol=42]}}
& 0.679 \\

08 &
\makecell[tl]{$\text{collide}(\text{shape=sphere},\text{shape=cube}) \to (\text{collide\_before\_half} \vee \text{entered\_then\_collided})$\\
{[mined-OR; conf=0.94, hits=1935, viol=121]}}
& 0.515 \\
\end{longtable}
\endgroup

\begingroup
\setlength{\tabcolsep}{3pt}
\renewcommand{\arraystretch}{1.08}
\tiny
\begin{longtable}{p{0.06\columnwidth}p{0.66\columnwidth}r}
\caption{Per-rule CLEVRER results for the \textbf{Neural Evaluator}. For readability, repeated predicate arguments within the same row are shown only on first occurrence.}
\label{tab:clevrer_neural_perrule_onecol}\\
\toprule
ID & Rule & AUROC \\
\midrule
\endfirsthead

\toprule
ID & Rule & AUROC \\
\midrule
\endhead

\midrule
\multicolumn{3}{r}{Continued on next page}
\endfoot

\bottomrule
\endlastfoot

01 &
\makecell[tl]{[Chain] $(\text{collide}(\text{shape=sphere},\text{color=brown}) \to \text{collide\_before\_half})$\\
$\wedge\;(\text{collide\_before\_half} \to \text{entered\_then\_collided})$}
& 0.895 \\

02 &
\makecell[tl]{[Chain] $(\text{collide}(\text{shape=cube},\text{color=cyan}) \to \text{collide\_before\_half})$\\
$\wedge\;(\text{collide\_before\_half} \to \text{entered\_then\_collided})$}
& 0.891 \\

03 &
\makecell[tl]{[Chain] $(\text{collide}(\text{shape=sphere},\text{color=green}) \to \text{entered\_then\_collided})$\\
$\wedge\;(\text{entered\_then\_collided} \to \text{collide\_before\_half})$}
& 0.883 \\

04 &
\makecell[tl]{[Chain] $(\text{collide}(\text{shape=cube},\text{color=blue}) \to \text{entered\_then\_collided})$\\
$\wedge\;(\text{entered\_then\_collided} \to \text{collide\_before\_half})$}
& 0.893 \\

05 &
\makecell[tl]{$\text{collide}(\text{shape=sphere},\text{shape=cube}) \to (\text{collide\_before\_half} \wedge \text{entered\_then\_collided})$\\
{[mined-AND; conf=0.51, hits=1041, viol=1015]}}
& 0.760 \\

06 &
\makecell[tl]{$\text{collide}(\text{shape=sphere},\text{shape=cylinder}) \to (\text{collide\_before\_half} \wedge \text{entered\_then\_collided})$\\
{[mined-AND; conf=0.49, hits=1148, viol=1212]}}
& 0.728 \\

07 &
\makecell[tl]{$\text{collide}(\text{shape=cube},\text{color=cyan}) \to (\text{collide\_before\_half} \vee \text{entered\_then\_collided})$\\
{[mined-OR; conf=0.95, hits=726, viol=42]}}
& 0.868 \\

08 &
\makecell[tl]{$\text{collide}(\text{shape=sphere},\text{shape=cube}) \to (\text{collide\_before\_half} \vee \text{entered\_then\_collided})$\\
{[mined-OR; conf=0.94, hits=1935, viol=121]}}
& 0.764 \\
\end{longtable}
\endgroup

\subsection{OpenImages -- Detailed Results Tables}

\begin{table*}[h]
\centering
\scriptsize
\setlength{\tabcolsep}{4pt}
\begin{tabular}{lrrrc@{\hspace{10pt}}lrrrc}
\toprule
Class & Prev. & ROC-AUC & AP & Acc.@0.5 &
Class & Prev. & ROC-AUC & AP & Acc.@0.5 \\
\midrule
Man           & 3734 & 0.931 & 0.691 & 0.885 &
Tower         & 143  & 0.994 & 0.662 & 0.976 \\
Tree          & 3611 & 0.932 & 0.772 & 0.911 &
Human nose    & 2612 & 0.952 & 0.742 & 0.891 \\
Human face    & 3124 & 0.950 & 0.785 & 0.884 &
Bicycle wheel & 245  & 0.993 & 0.806 & 0.942 \\
Person        & 7144 & 0.827 & 0.637 & 0.771 &
Glasses       & 258  & 0.970 & 0.548 & 0.935 \\
Woman         & 2775 & 0.942 & 0.720 & 0.889 &
Dress         & 710  & 0.961 & 0.552 & 0.922 \\
Footwear      & 1907 & 0.936 & 0.570 & 0.877 &
Vehicle       & 1545 & 0.639 & 0.118 & 0.800 \\
Window        & 904  & 0.937 & 0.414 & 0.868 &
Bird          & 593  & 0.985 & 0.801 & 0.941 \\
\bottomrule
\end{tabular}
\caption{OpenImages leaf-bank evaluation on the validation split. ``Prev.'' denotes the number of positive validation images for the class.}
\label{tab:oi_leaf_perclass_compact}
\end{table*}

\begin{table}[h]
\centering
\small
\setlength{\tabcolsep}{6pt}
\begin{tabular}{ll}
\toprule
Not-top49 class & Ancestors in Top-49 \\
\midrule
Truck       & Land vehicle, Vehicle \\
Van         & Land vehicle, Vehicle, Car \\
Motorcycle  & Land vehicle, Vehicle \\
Helicopter  & Vehicle \\
Watercraft  & Vehicle \\
Cart        & Land vehicle, Vehicle \\
Bus         & Land vehicle, Vehicle \\
Train       & Land vehicle, Vehicle \\
Tank        & Land vehicle, Vehicle, Building \\
Wheelchair  & Land vehicle, Vehicle \\
Taxi        & Land vehicle, Vehicle \\
Limousine   & Land vehicle, Vehicle, Car \\
Gondola     & Boat, Vehicle \\
Ambulance   & Land vehicle, Vehicle \\
Segway      & Land vehicle, Vehicle \\
Barge       & Boat, Vehicle \\
Golf cart   & Land vehicle, Vehicle \\
Snowmobile  & Land vehicle, Vehicle \\
Seat belt   & Land vehicle, Vehicle \\
\bottomrule
\end{tabular}
\caption{Examples of classes outside the Top-49 whose ancestors are nevertheless present within the Top-49 set. This illustrates how hierarchical closure can introduce parent-level signals into evaluation even when the corresponding fine-grained child class is not itself included in the selected label vocabulary.}
\label{tab:oi_nontop49_ancestors}
\end{table}

\begin{table*}[h]
\centering
\scriptsize
\setlength{\tabcolsep}{4pt}
\begin{tabular}{llcc}
\toprule
ID & Rule & Indep. Events & Neural Eval. \\
\midrule
001 & Tableware $\to$ Bottle & 0.891 & 0.954 \\
002 & Furniture $\to$ Chair & 0.830 & 0.934 \\
003 & Furniture $\to$ Table & 0.819 & 0.934 \\
004 & Building $\to$ Skyscraper & 0.869 & 0.911 \\
005 & Building $\to$ Tower & 0.867 & 0.908 \\
006 & Building $\to$ House & 0.843 & 0.887 \\
007 & Bicycle $\to$ Bicycle wheel & 0.668 & 0.925 \\
008 & Tree $\to$ Palm tree & 0.902 & 0.940 \\
009 & Vehicle $\to$ Land vehicle & 0.837 & 0.798 \\
010 & Land vehicle $\to$ Bicycle & 0.788 & 0.986 \\
011 & Land vehicle $\to$ Car & 0.722 & 0.829 \\
012 & Bicycle wheel $\to$ Bicycle & nan & nan \\
013 & Furniture $\to$ (Chair $\wedge$ Table) & 0.854 & 0.938 \\
014 & Building $\to$ (Skyscraper $\wedge$ Tower) & 0.873 & 0.877 \\
015 & Building $\to$ (Tower $\wedge$ House) & 0.877 & 0.888 \\
016 & Land vehicle $\to$ (Bicycle $\wedge$ Car) & 0.765 & 0.989 \\
\bottomrule
\end{tabular}
\caption{Per-rule OpenImages results (ROC-AUC) comparing the Independent-Events probabilistic baseline and the full Neural Evaluator. Rule~012 has undefined AUROC due to degenerate evaluation labels.}
\label{tab:oi_indep_vs_neural_perrule}
\end{table*}

\subsection{VidOR -- Detailed Results Tables}

\begingroup
\setlength{\tabcolsep}{3pt}
\renewcommand{\arraystretch}{1.03}
\tiny
\begin{longtable}{p{0.56\columnwidth}rrrr}
\caption{VidOR leaf-bank evaluation on the validation split. ``Prev.'' denotes the number of positive validation videos for the class.}
\label{tab:vidor_leaf_perclass_onecol}\\
\toprule
Class & Prev. & ROC-AUC & AP & Acc.@thr \\
\midrule
\endfirsthead

\toprule
Class & Prev. & ROC-AUC & AP & Acc.@thr \\
\midrule
\endhead

\midrule
\multicolumn{5}{r}{Continued on next page}
\endfoot

\bottomrule
\endlastfoot

\multicolumn{5}{l}{\textbf{Object leaves}}\\
\midrule
obj:adult & 607 & 0.483 & 0.716 & 0.727 \\
obj:child & 299 & 0.625 & 0.442 & 0.640 \\
obj:toy & 115 & 0.688 & 0.219 & 0.862 \\
obj:baby & 122 & 0.697 & 0.245 & 0.854 \\
obj:chair & 65 & 0.641 & 0.119 & 0.922 \\
obj:dog & 73 & 0.622 & 0.122 & 0.913 \\
obj:car & 56 & 0.769 & 0.173 & 0.933 \\
\end{longtable}
\endgroup

\begingroup
\setlength{\tabcolsep}{3pt}
\renewcommand{\arraystretch}{1.05}
\tiny
\begin{longtable}{p{0.06\columnwidth}p{0.70\columnwidth}r}
\caption{Per-rule VidOR results for the Independent-Events Probabilistic evaluator.}
\label{tab:vidor_indep_perrule_onecol}\\
\toprule
ID & Rule & AUROC \\
\midrule
\endfirsthead

\toprule
ID & Rule & AUROC \\
\midrule
\endhead

\midrule
\multicolumn{3}{r}{Continued on next page}
\endfoot

\bottomrule
\endlastfoot

001 & \makecell[tl]{$\text{rel:adult-watch-adult} \to \text{rel:adult-next\_to-adult}$\\
{[mined rev; sup=1111, conf=0.792, lift=2.68]}} & 0.557 \\
002 & \makecell[tl]{$\text{rel:adult-watch-adult} \to \text{rel:adult-in\_front\_of-adult}$\\
{[mined rev; sup=1155, conf=0.823, lift=2.74]}} & 0.714 \\
003 & \makecell[tl]{$\text{rel:adult-away-adult} \to \text{rel:adult-next\_to-adult}$\\
{[mined rev; sup=540, conf=0.770, lift=2.60]}} & 0.477 \\
004 & \makecell[tl]{$\text{rel:adult-towards-adult} \to \text{rel:adult-next\_to-adult}$\\
{[mined rev; sup=507, conf=0.791, lift=2.67]}} & 0.650 \\
005 & \makecell[tl]{$\text{rel:adult-behind-adult} \to \text{rel:adult-in\_front\_of-adult}$\\
{[mined rev; sup=1203, conf=0.914, lift=3.04]}} & 0.451 \\
006 & \makecell[tl]{$\text{rel:adult-away-adult} \to \text{rel:adult-in\_front\_of-adult}$\\
{[mined rev; sup=593, conf=0.846, lift=2.81]}} & 0.633 \\
007 & \makecell[tl]{$\text{rel:adult-watch-child} \to \text{rel:child-in\_front\_of-adult}$\\
{[mined rev; sup=601, conf=0.854, lift=5.51]}} & 0.563 \\
008 & \makecell[tl]{$\text{rel:child-watch-adult} \to \text{rel:adult-in\_front\_of-child}$\\
{[mined rev; sup=346, conf=0.783, lift=6.89]}} & 0.297 \\
\end{longtable}
\endgroup

\begingroup
\setlength{\tabcolsep}{3pt}
\renewcommand{\arraystretch}{1.05}
\tiny
\begin{longtable}{p{0.06\columnwidth}p{0.70\columnwidth}r}
\caption{Per-rule VidOR results for the Neural Evaluator.}
\label{tab:vidor_neural_perrule_onecol}\\
\toprule
ID & Rule & AUROC \\
\midrule
\endfirsthead

\toprule
ID & Rule & AUROC \\
\midrule
\endhead

\midrule
\multicolumn{3}{r}{Continued on next page}
\endfoot

\bottomrule
\endlastfoot

001 & \makecell[tl]{$\text{rel:adult-watch-adult} \to \text{rel:adult-next\_to-adult}$\\
{[mined rev; sup=1111, conf=0.792, lift=2.68]}} & 0.653 \\
002 & \makecell[tl]{$\text{rel:adult-watch-adult} \to \text{rel:adult-in\_front\_of-adult}$\\
{[mined rev; sup=1155, conf=0.823, lift=2.74]}} & 0.771 \\
003 & \makecell[tl]{$\text{rel:adult-away-adult} \to \text{rel:adult-next\_to-adult}$\\
{[mined rev; sup=540, conf=0.770, lift=2.60]}} & 0.677 \\
004 & \makecell[tl]{$\text{rel:adult-towards-adult} \to \text{rel:adult-next\_to-adult}$\\
{[mined rev; sup=507, conf=0.791, lift=2.67]}} & 0.626 \\
005 & \makecell[tl]{$\text{rel:adult-behind-adult} \to \text{rel:adult-in\_front\_of-adult}$\\
{[mined rev; sup=1203, conf=0.914, lift=3.04]}} & 0.674 \\
006 & \makecell[tl]{$\text{rel:adult-away-adult} \to \text{rel:adult-in\_front\_of-adult}$\\
{[mined rev; sup=593, conf=0.846, lift=2.81]}} & 0.696 \\
007 & \makecell[tl]{$\text{rel:adult-watch-child} \to \text{rel:child-in\_front\_of-adult}$\\
{[mined rev; sup=601, conf=0.854, lift=5.51]}} & 0.778 \\
008 & \makecell[tl]{$\text{rel:child-watch-adult} \to \text{rel:adult-in\_front\_of-child}$\\
{[mined rev; sup=346, conf=0.783, lift=6.89]}} & 0.505 \\
\end{longtable}
\endgroup

\section{Independent events formulas}\label{indev}
Assume leaf events are \textbf{independent} \emph{given} $x$, then we compute a soft satisfaction probability $P(\varphi)$ by recursion using these independent events formulas:

\begin{align}
P(\neg A) &= 1 - p_A,\\
P(A\wedge B) &= p_A\,p_B,\\
P(A\vee B) &= p_A + p_B - p_A p_B,\\
P(A\Rightarrow B) &= 1 - P(A\wedge \neg B) = 1 - p_A(1-p_B),\\
P(A\Leftrightarrow B) &= P(A\wedge B) + P(\neg A\wedge \neg B)
= p_A p_B + (1-p_A)(1-p_B).
\end{align}

\section{Neural Algebra of Classifiers (NAC) and similar methods, and relation to our evaluator (extended discussion).} 
Neural Algebra of Classifiers (NAC; \cite{SantaCruz2018NAC}) (other similar methods are \citep{Misra_2017_CVPR,Nagarajan_2018_ECCV,Yang_2020_CVPR,Li_2021_WACV})) is the closest conceptual precedent to our work in that it learns neural modules intended to implement Boolean connectives and composes them along an expression tree. However, NAC composes classifier parameters (e.g., weight vectors for primitive concept classifiers) to synthesize a new classifier for a composed expression, and is trained primarily with expression-level supervision (labels for the whole composed concept). In contrast, our evaluator composes sample-level evidence through an explicit rule DAG to output clause-and rule-satisfaction probabilities, is trained with internal-node logical supervision obtained by hard propagation from ground-truth concept labels, and uses chimera operand mixing to discourage shortcut solutions and enforce operator-level compositionality locally. Finally, we introduce lineage-aware subtree caching (keyed by symbolic structure and encoder fingerprint) to reuse learned submodules safely across large rule sets and across runs, addressing scalability and representation drift in a way orthogonal to NAC’s global operator design. 

The difference is that NAC can learn to evaluate, in a fully nonzero\_support-supervised way, a given set of rules, and then compose to evaluate on new rules not used during training; on the other hand, in our approach one can re-use only sub-rules of a bigger rule that was learned to be evaluated, but both rules didn't need any nonzero support in the training set to be learned. Thus, the strategy is to simply train our system with a very big rule containing as sub-rules all of the rules of interest, with the limiting factor only being the trade-off between the size of the total rule vs. training time, but not the nature of the training dataset itself w.r.t. to its support of the rules during supervision; nevertheless, at test time, given any sub-rule, the forward pass is simply an inference along the corresponding graph of classifiers.

One may speculate that, in NAC, the learned signal (from a fixed set of supervised training rules) will degrade at some point in a long chain of compositions, since the novel rules may be reactive to intricate correlations in uncertainty that were simply not captured by the initial set of training rules. By contrast, our local training signal at every sub-depth in the graph ensures that this should not be the case.

\section{Qualitative sanity-check experiment: MNIST contradiction rule}
\label{sec:qual_mnist_contradiction}

\paragraph{Goal.}
This experiment is designed as a \emph{qualitative} diagnostic of the core ideas (rule-graphs, negation handling, learned subtree gates, and caching), not as a benchmark result. We deliberately choose a rule whose truth value is \emph{identically false} for all inputs. The only acceptable behavior is that the learned root satisfaction probability stays near zero across the dataset, without spuriously correlating with visual styles or digit morphology.

\paragraph{Setup: concepts and leaf bank.}
We use MNIST digits as a simple controlled perceptual domain. The concept vocabulary is the 10-way one-hot digit identity:
\[
y\in\{0,1\}^{10},\qquad y_d = \mathbb{I}[\text{digit}=d],\ \ d\in\{0,\dots,9\}.
\]
We train a lightweight convolutional leaf concept bank with a shared encoder and $10$ sigmoid heads using multi-label BCE (even though labels are one-hot). This yields (i) encoder features $z=E_\phi(x)$ and (ii) leaf probabilities $p(x)\in(0,1)^{10}$.

\paragraph{Rule graph: \(A \Leftrightarrow \neg A\).}
Fix a target digit $n\in\{0,\dots,9\}$ and define the atomic proposition
\[
A \equiv (\text{digit}=n).
\]
We compile a 3-node DGL DAG with two leaf nodes referencing the \emph{same} concept ID (digit $n$) and a single root IFF node:
\[
\text{root} \;=\; A \Leftrightarrow \neg A.
\]
Concretely, the graph has edges (leaf $\to$ root) with negation flags $(+1,-1)$ so that the second child is negated. Under exact Boolean semantics, this formula is false for every input:
\[
\forall x,\quad (A(x)\Leftrightarrow \neg A(x)) = 0.
\]
Therefore, the ground-truth root label is constant: $t_{\text{root}}(x)=0$ for all $x$.

\paragraph{Training: single-level gate learning.}
Since the rule has depth 1, training reduces to learning \emph{one} subtree gate at the root. We use the standard level-wise training procedure:
\begin{enumerate}
  \item For each mini-batch, compute the hard truths for all nodes by bottom-up propagation from concept labels (so the root target is always 0).
  \item Initialize both leaves with the encoder feature vector $z$ (in this construction both leaves point to the same concept and thus carry the same base evidence), and pass the two child features plus negation flags into the root gate.
  \item Optimize root-gate BCE loss for a small number of epochs and store the trained gate in the subtree cache keyed by the rule structure and encoder fingerprint.
\end{enumerate}

A useful qualitative difference emerges when one sorts the test images of a fixed digit class by the score assigned to this contradiction rule. In this special sanity-check experiment, we use the rule output itself as the anomaly score for visualization, rather than the usual $1-p$ transformation used for rule-satisfaction scores elsewhere in the paper. This makes the comparison especially revealing. Under the SEM variant, training sees only real normal same-image samples, which is just the full dataset in this case since no real image in it has labels that would make the truth value of this rule (which is an impossible logical assertion) nonzero. As a result, SEM receives no explicit supervisory signal that would force it to organize within-class variation according to visual abnormality; any such ordering can only arise indirectly from residual classifier uncertainty. Chimera training, by contrast, augments the same normal data with synthetic contradictory examples that are impossible in real life but are semantically `true' for the rule. This provides an explicit counterfactual signal for what ``abnormal'' should look like at the rule level. Qualitatively, this changes the ranking behavior: although all test images shown in the figure have the same class label, the samples assigned the largest contradiction scores are visually more distorted, less prototypical, or harder to parse than those assigned the smallest scores. The chimera-trained evaluator produces a noticeably cleaner progression from normal-looking digits (left) to abnormal-looking digits (right) than SEM, suggesting that the synthetic contradictory supervision helps the model detect within-class visual abnormality rather than merely reproducing the nominal dataset label.

\begin{figure}
    \centering
    \includegraphics[width=1\linewidth]{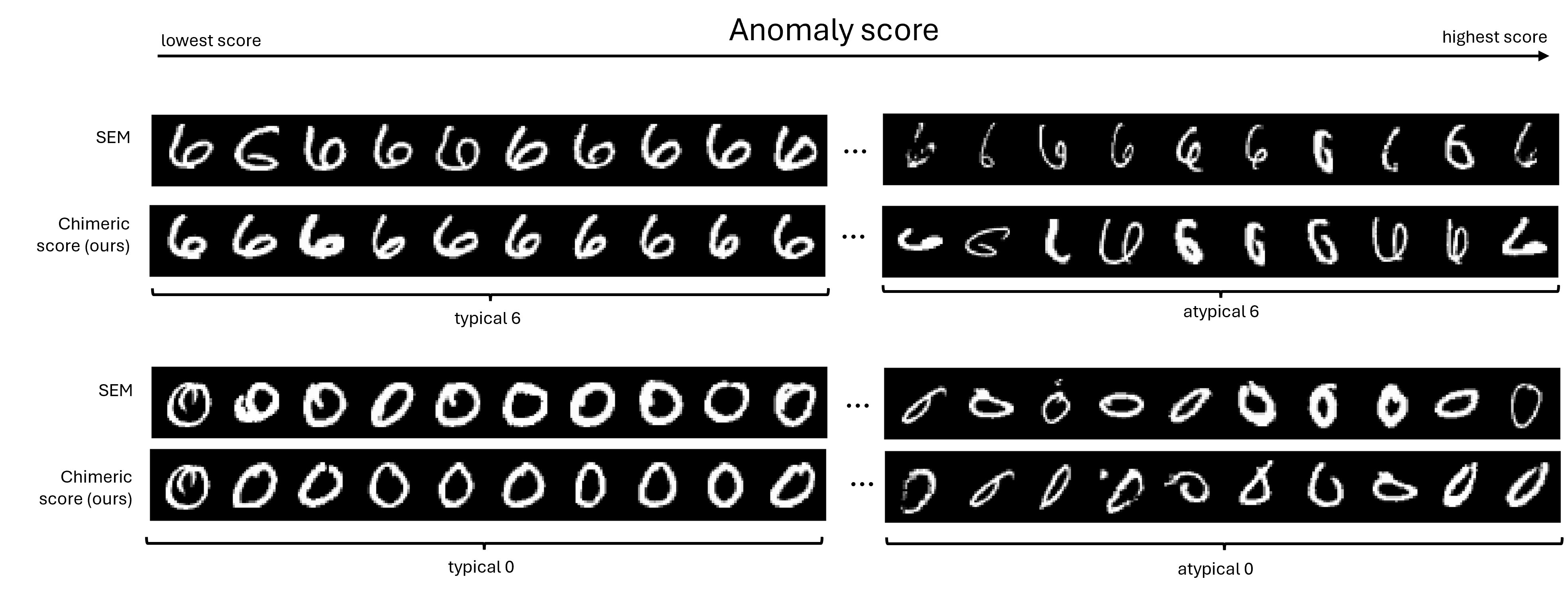}
    \caption{\textbf{Qualitative visualization of results for the MNIST contradiction rule} ${A \Leftrightarrow \neg A}$. In this experiment, the anomaly score is the model output for the contradiction rule itself (not $1-p$). SEM is trained only on real, same-image normal samples, whereas chimera training additionally introduces synthetic contradictory examples that cannot occur in real data. For a fixed test class, we sort samples by anomaly score and show the 10 smallest on the left and the 10 largest on the right. Although all shown images share the same dataset label, the high-score samples are visually more distorted and less prototypical. Chimera produces a visibly cleaner separation, tending to place more normal-looking digits on the left and more abnormal-looking ones on the right, which suggests that the synthetic contradictory supervision helps the evaluator detect within-class visual abnormality more effectively.}
    \label{fig:mnist}
\end{figure}

\section{Qualitative results in images}
\label{sec:qual}

\begin{figure}[!h]
    \centering
    \includegraphics[width=1\linewidth]{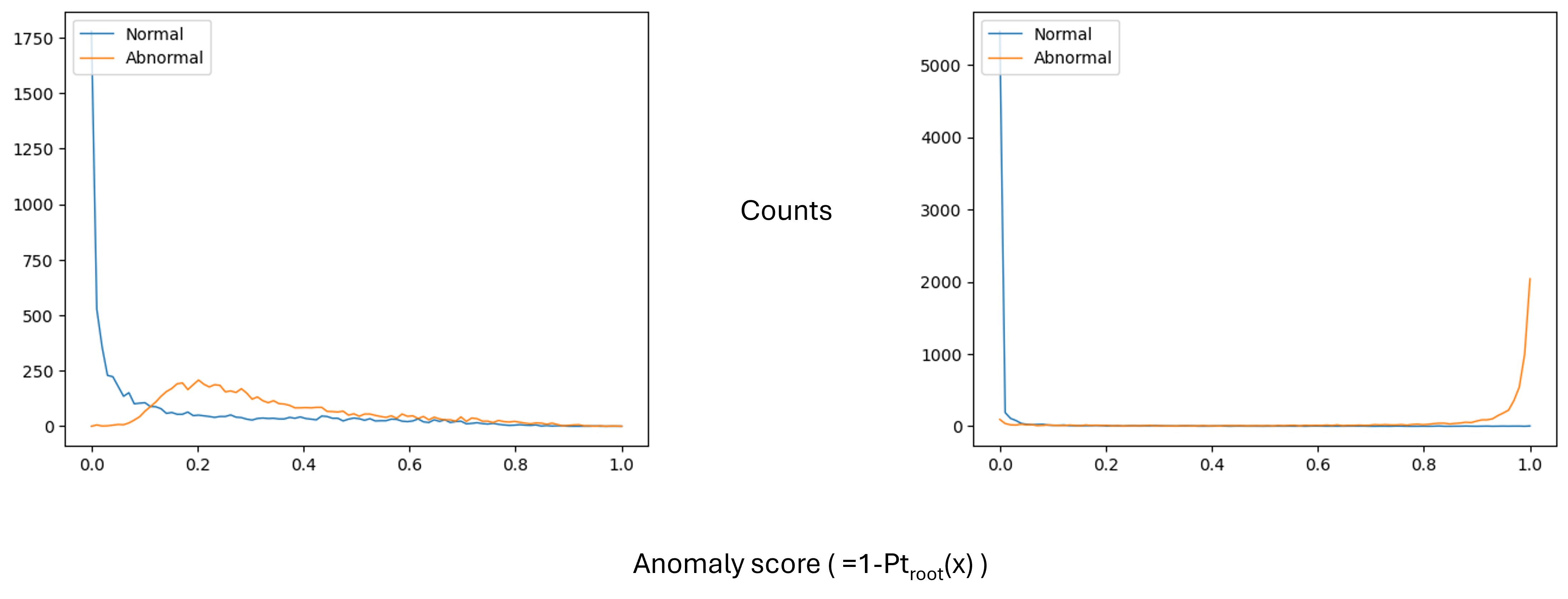}
    \caption{Anomaly score histograms corresponding to the results of the experiments in the figures below (OpenImages - Rule: Land vehicle $\to$ (Bicycle $\wedge$ Car)), Figs. \ref{fig:oir1}-\ref{fig:oir2}. \textbf{Left}, Indep.Events; \textbf{right}, Neural Evaluator.}
    \label{fig:oih}
\end{figure}

\begin{figure}[!h]\label{fig:oir1}
    \centering
    \includegraphics[width=1\linewidth]{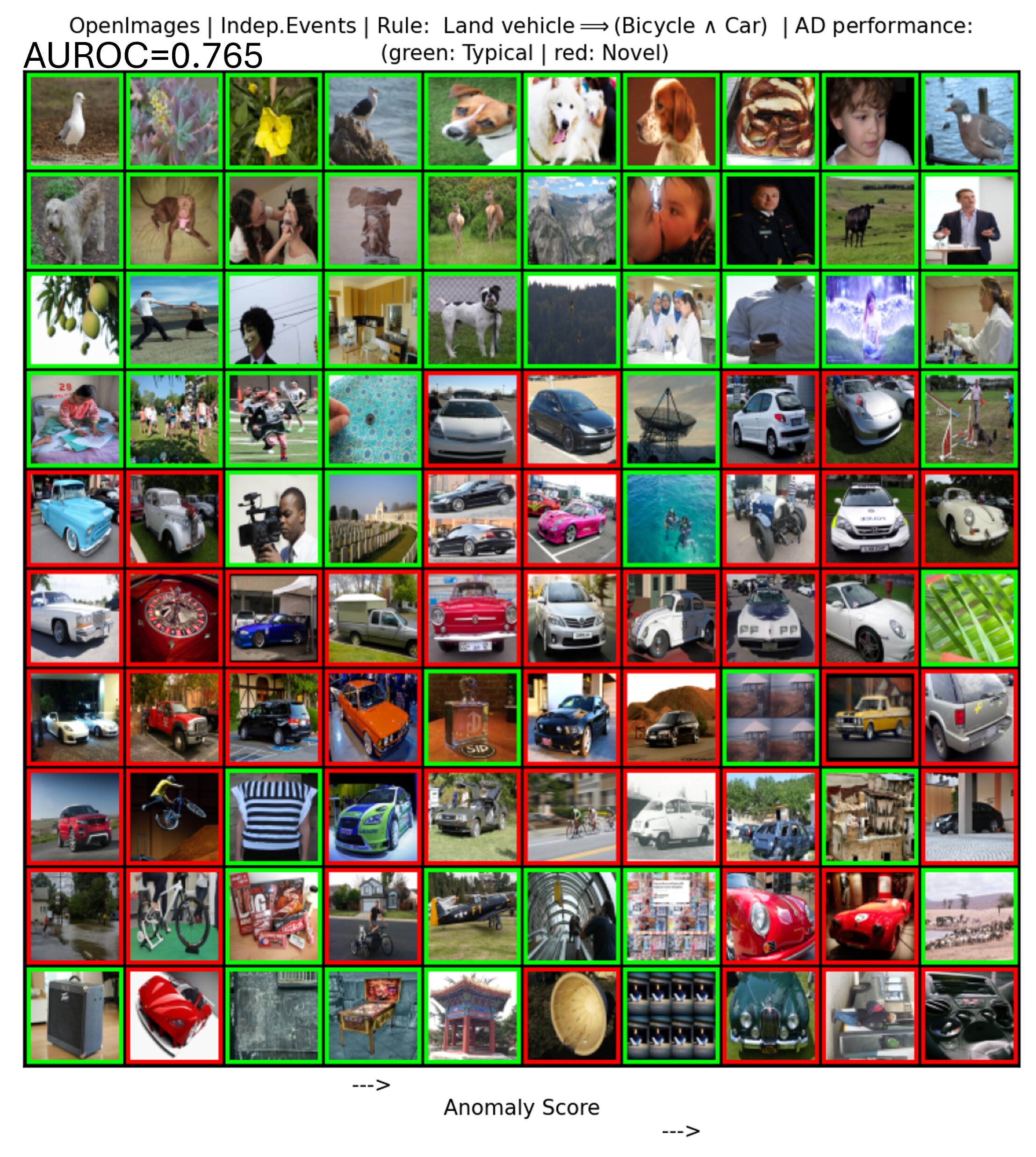}
    \caption{Showing the images from a random but \textit{balanced} subset of the test set (i.e., same number of normal and abnormal images), with the index sorted by the anomaly score, $s_r(x) = 1 - \hat{t}^{(r)}_{\mathrm{root}}(x)$, from low at the top left to high at the bottom right. Only displaying one every 10 images, starting from index 0. A perfect detection would show the top half of the \textit{total} panel as normal (green framing) and the anomalies (red framing) at the bottom half.}
\end{figure}

\begin{figure}[!h]
    \centering
    \includegraphics[width=1\linewidth]{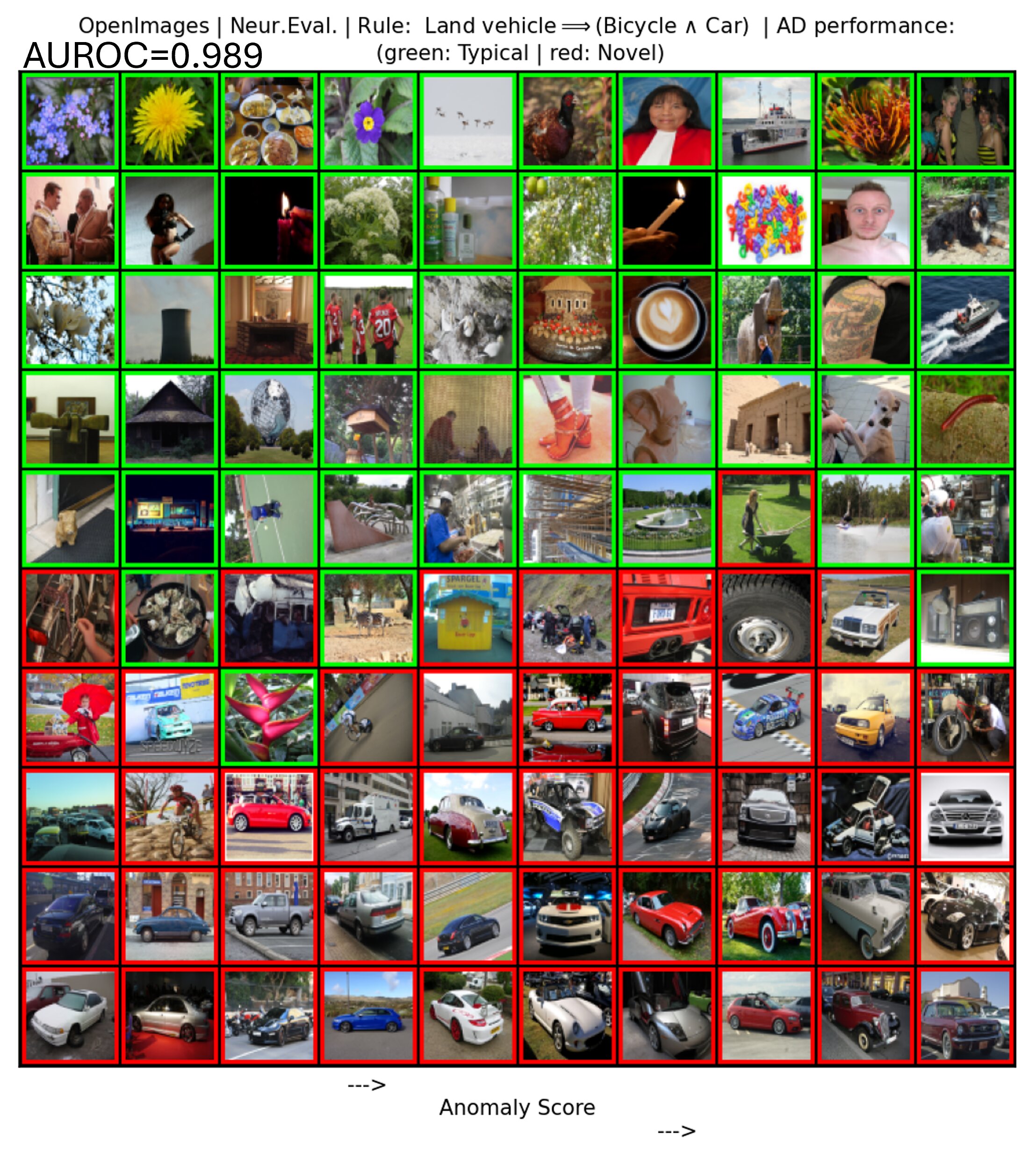}
    \caption{Idem previous figure.}
    \label{fig:oir2}
\end{figure}

\clearpage
\newpage

\section{Training protocol and architecture details}

We use the implementation on the VidOR dataset as example. The implementation on the other datasets uses the same scripts with minimal adaptations to fit them.

\subsection{Experimental Reproducibility Details}
\label{app:reproducibility}

\paragraph{Codebase and implementation.}
All experiments use a PyTorch/DGL implementation of a feature-aware neural rule evaluator. 
Each logical rule is represented as a directed acyclic graph whose leaves are semantic concepts and whose internal nodes are Boolean operators. 
The evaluator is trained bottom-up from leaves to root: each internal gate maps child features, together with edge-level negation flags, to a parent feature and a parent truth probability. 
The implementation uses DGL graph traversal for the rule structure, PyTorch modules for the leaf bank and subtree gates, and scikit-learn for AUROC/AP evaluation.

\paragraph{Rule evaluator and caching.}
For each rule graph, internal modules are trained level-wise by node depth. 
The implementation uses lineage-aware subtree caching: cache keys include the exact symbolic subtree, edge order/negations, gate architecture tag, feature dimensionality, and a fingerprint of the upstream encoder. 
This prevents accidental reuse of subtree gates when the feature-producing encoder changes. 
At inference, the trained or cached gates are applied level-by-level until the root truth probability is obtained.

\subsection{Dataset Construction and Splits}
\label{app:data}

\paragraph{VidOR data layout.}
The VidOR experiments assume the following directory layout:
\[
\texttt{<vidor\_root>/annotations/\{train,val\}/*.json},\] 
\[
\texttt{<vidor\_root>/videos/<video\_path>}.
\]
If the annotation does not directly provide a resolvable video path, the code falls back to a video identifier lookup. 
The train split is used for concept mining, rule mining, leaf-bank training, rule-gate training, optional calibration, and optional learned aggregation. 
The validation split is used for final anomaly scoring and reported AUROC.

\paragraph{Semantic leaf vocabulary.}
The VidOR leaf vocabulary consists of object atoms and relation atoms:
\[
\mathrm{obj}:\langle \mathrm{category}\rangle,
\qquad
\mathrm{rel}:\langle \mathrm{subject}\rangle
\text{-}
\langle \mathrm{predicate}\rangle
\text{-}
\langle \mathrm{object}\rangle .
\]
The object vocabulary is selected from the most frequent training categories, and the relation vocabulary is selected from the most frequent training subject--predicate--object triples after applying the minimum relation-support threshold. 
In the reported configuration, the default budgets are
\[
K_{\mathrm{obj}}=40,\qquad K_{\mathrm{rel}}=200,\qquad 
\mathrm{min\_rel\_support}=20,
\]
unless otherwise stated in the experiment table.

\paragraph{Video preprocessing.}
Each video is represented by uniformly sampled frames. 
The default VidOR configuration uses
\[
T=8 \quad \text{frames per clip}, \qquad 224\times224 \quad \text{spatial resolution}.
\]
At training time, optional augmentation consists of resizing, color jitter, random affine perturbations, tensor conversion, and random erasing. 
At validation time, only resizing and tensor conversion are applied.

\subsection{Model Architecture}
\label{app:architecture}

\paragraph{Leaf bank.}
The VidOR leaf bank is a multi-label video classifier. 
For each clip, a frame-level CNN backbone is applied to the sampled frames, the resulting frame embeddings are projected to a feature dimension \(F\), and the projected frame embeddings are mean-pooled across time. 
The pooled feature is fed into \(K\) independent linear heads, one per semantic leaf. 
The default backbone is ImageNet-pretrained ResNet-18, with a lightweight convolutional alternative available. 
The default feature dimension is
\[
F=256.
\]

\paragraph{Subtree gate.}
Each internal node of arity \(a\) uses a local gate
\[
g_{\theta}: \mathbb{R}^{a(F+1)} \rightarrow \mathbb{R}^{F}\times(0,1),
\]
where the input concatenates each child feature with a scalar edge-negation flag. 
The gate is an MLP with ReLU nonlinearities followed by a sigmoid truth head:
\[
(h_v,\hat{p}_v)=g_{\theta_v}
\left(
[h_{c_1},b_1,\ldots,h_{c_a},b_a]
\right).
\]
The output \(h_v\) is used as the parent feature for higher nodes, and \(\hat{p}_v\) is the predicted truth probability for the subformula rooted at \(v\).

\subsection{Training Protocol and Hyperparameters}
\label{app:training}

\paragraph{Optimization.}
The leaf bank is trained with binary cross-entropy with logits. 
When class-count statistics are enabled, positive-class weights are computed from training-set prevalence. 
The default optimizer for the leaf bank is Adam with learning rate
\[
\eta_{\mathrm{leaf}}=10^{-3},
\]
weight decay \(0\) unless otherwise stated, and default training duration of \(3\) epochs.

\paragraph{Level-wise evaluator training.}
For every rule graph, subtree gates are trained bottom-up by depth. 
At each level, hard truth targets for all graph nodes are computed from ground-truth leaf labels using exact Boolean propagation. 
The gate at each internal node is optimized with binary cross-entropy against its node-level Boolean truth target. 
The default evaluator training hyperparameters are:
\[
\eta_{\mathrm{level}}=10^{-3},\qquad
\mathrm{epochs\_level}=2,\qquad
\mathrm{batch\_train}=64.
\]

\paragraph{Chimera-only operand training.}
All reported runs use (except in the ablations that use only the normal class during training)
\[
\texttt{--negatives chimeras\_only}.
\]
In this mode, the operands of each binary gate are constructed from different samples in the mini-batch. 
For a batch permutation \(\pi\) with \(\pi(i)\neq i\), the left operand is taken from sample \(i\) and the right operand from sample \(\pi(i)\). 
The target is computed by applying the Boolean operator to the corresponding hard child truth values:
\[
t^{\mathrm{chim}}_i
=
\operatorname{op}\!\left(t_{\ell}(y_i),t_r(y_{\pi(i)})\right).
\]
Both positive and negative chimera cases are used, depending on whether the operator evaluates to true or false.

\paragraph{Temperature calibration.}
When enabled, post-hoc temperature scaling is fit on the training split using binary cross-entropy on the leaf logits. 
The learned scalar temperature is saved and reloaded at evaluation time.

\paragraph{Default run configuration.}
Unless explicitly overridden in the experiment table, the VidOR runs use:
\[
\begin{array}{ll}
\text{Backbone} & \text{ResNet-18} \\
\text{Feature dimension} & 256 \\
\text{Frames per clip} & 8 \\
\text{Resize} & 224\times 224 \\
\text{Leaf epochs} & 3 \\
\text{Rule-gate epochs per level} & 2 \\
\text{Leaf learning rate} & 10^{-3} \\
\text{Rule-gate learning rate} & 10^{-3} \\
\text{Batch size, train/eval} & 64/64 \\
\text{Rule aggregation} & \texttt{min} \\
\text{Implication gate threshold} & 0.0 \\
\text{Random seed} & 123 .
\end{array}
\]

\subsection{Evaluation Protocol}
\label{app:evaluation}

\paragraph{Leaf-level evaluation.}
For the leaf bank, we report per-class ROC-AUC, average precision, and accuracy at threshold \(0.5\) on the validation split. 
Macro summaries are computed over valid classes, ignoring degenerate classes for which ROC-AUC or AP is undefined.

\paragraph{Rule-level scores.}
For each rule \(r\), the evaluator produces a predicted rule-satisfaction probability
\[
\hat{p}_r(x)\in(0,1).
\]
The per-rule violation score is
\[
v_r(x)=1-\hat{p}_r(x).
\]

\paragraph{AUROC computation.}
The main anomaly metric is AUROC against the pseudo-anomaly label
\[
Y_{\mathrm{anom}}(x)=1-\min_r T_r(y),
\]
where \(T_r(y)\in\{0,1\}\) is the exact Boolean truth value of rule \(r\) under ground-truth leaves. 
Per-rule AUROC is also computed by comparing \(1-\hat{p}_r(x)\) against \(1-T_r(y)\). 
If the validation labels are all normal or all anomalous under the pseudo-ground-truth, AUROC is reported as undefined rather than forced to a numeric value.

\subsection{Compute and Software Environment}
\label{app:compute}

\paragraph{Software dependencies.}
The experiments require Python with PyTorch, torchvision, DGL, OpenCV, Pillow, scikit-learn, NumPy, and tqdm. 
The code selects CUDA automatically when available and otherwise falls back to CPU.

\paragraph{Hardware.}
The experiments were run on a NVIDIA H100 GPU.

\paragraph{Runtime controls.}
The most important runtime controls are the number of sampled frames, image resize, batch size, number of workers, number of leaf epochs, number of evaluator epochs per level, and the number of retained rules. 
The code supports \texttt{--train\_frac} for controlled training-set subsampling and \texttt{--train\_missing\_only} for reusing existing cached gates when applicable.

\subsection{Randomness and Determinism}
\label{app:randomness}

The code sets the Python \texttt{random} seed and the PyTorch CPU seed, and also sets the CUDA seed when CUDA is available. 
The default seed is
\[
123
\]
for the VidOR experiments. 
Remaining nondeterminism can arise from GPU kernels, data-loader worker scheduling, video decoding, and randomized data augmentation.

\subsection{Data, Licenses, and Ethics}
\label{app:data_ethics}

\paragraph{Dataset access.}
We use existing third-party datasets and do not introduce a new dataset. 
The VidOR files are expected to be obtained from the official dataset distribution and arranged in the directory structure described above.

\paragraph{Annotations and labels.}
The anomaly labels used in our experiments are pseudo-labels derived from logical inconsistency under retained semantic rules, not human judgments of abnormality. 
Therefore, the reported anomaly-detection results should be interpreted as rule-consistency detection under a selected concept vocabulary and selected rule set.

\paragraph{Privacy and human subjects.}
The experiments use pre-existing public video annotations and do not involve newly collected human-subject data. 
No attempt is made to identify individuals. 
If any dataset split contains people, the analysis is restricted to the dataset-provided object and relation categories.

\paragraph{Limitations and potential misuse.}
The method can flag violations of explicit rules, but it inherits errors from the leaf bank, biases from dataset annotations, and biases from mined rule selection. 
A high anomaly score should therefore be interpreted as evidence of semantic inconsistency relative to the selected rule set, not as a general-purpose safety or surveillance judgment. 
The system should not be deployed for consequential decisions without validating the rule set, concept vocabulary, calibration, and false-positive/false-negative behavior in the target domain.

\clearpage
\newpage

\end{document}